\documentclass{article}

\usepackage{iclr2026_conference,times}
\iclrfinalcopy
\newtheorem{condition}{Condition}
\usepackage{amsmath,amssymb,amsfonts}
\usepackage{booktabs,multirow,colortbl}
\usepackage{graphicx}
\usepackage{enumitem}
\usepackage{xcolor}
\usepackage{microtype}
\usepackage[colorlinks=true,linkcolor=blue,citecolor=blue,urlcolor=blue]{hyperref}
\usepackage{tikz}
\usepackage{algorithm}
\usepackage{algpseudocode}
\usetikzlibrary{positioning,arrows.meta,calc,fit,backgrounds,decorations.pathreplacing}
\usepackage{pifont}
\usepackage{pgfplots}
\pgfplotsset{compat=1.16}
\usepgfplotslibrary{fillbetween}
\setlist{itemsep=2pt,topsep=2pt}

\newcommand{\method}{Transfer-Selective Replay}
\newcommand{\methodshort}{TSR}

\title{Rethinking Transfer in Continual Learning:\\A Replay-Based Realisation}

\author{Yang Meng, \ Zhenya Liu, \ Zhuokai Zhao$^\dagger$, \ Yuxin Chen$^\dagger$ \\
Department of Computer Science, University of Chicago \\
\texttt{ymeng3@uchicago.edu} \\
$^\dagger$\,Co-last authors.
}
\date{}

\begin{document}
\maketitle
{\let\thefootnote\relax\footnotetext{Code is available at
\url{https://github.com/ymeng3/transfer-selective-replay}.}}

\begin{abstract}
Continual learning studies how deployed language models can continually acquire new tasks without expensive retraining from scratch.  Existing methods, whether
rehearsal-based (replaying stored past data) or rehearsal-free
(regularising or isolating parameters), overwhelmingly target
one objective: preventing catastrophic forgetting.  Forward
transfer --- the past helping the future --- has meanwhile been
pursued almost exclusively through parameter reuse, with no
explicit account of when transfer should be expected at all.
We begin one step earlier: before designing a transfer
mechanism, we ask when transfer should exist at all.  We answer
with a framework of three measurable conditions: the target task must leave room for improvement beyond its own limited supervision, transferable information must survive continued optimisation, and replay must come from compatible previous tasks.  We instantiate this view as \method{} (\methodshort{}), which selects replay data predicted to benefit the incoming task rather than replaying past examples indiscriminately.  Selection is guided by a zero-training task signature, while distillation preserves stability on previous tasks. Under the standard
continual-learning protocol in the low-budget regime,
\methodshort{} consistently improves forward transfer while maintaining stability, outperforming existing replay
baselines across heterogeneous and homogeneous task streams.
More broadly, the results argue for treating transfer as a
first-class objective of continual learning, to be understood
before it is engineered.
\end{abstract}

\section{Introduction}\label{sec:intro}

Large language models are trained once, at enormous cost, and
then deployed into a world that does not stand still: new
domains, new formats, and new skills keep arriving after
release~\citep{wang2023trace,wang2024survey,kalyan2025curll,clbench2026}.  Retraining from
scratch for every addition is prohibitive, so a deployed model
must learn tasks \emph{sequentially} --- the setting of
continual learning~\citep{delange2021survey,keliu2022survey,inca2025,opr2026}.
Its defining failure mode has been known for decades: \emph{catastrophic forgetting}, where optimising for
the new task overwrites what earlier tasks
built~\citep{kirkpatrick2017ewc,lopezpaz2017gem,zheng2025spurious}.  The field
frames the resulting tension as a \emph{stability--plasticity}
trade-off: stability preserves the past, plasticity lets the
model keep learning, and improving either naively costs the
other~\citep{delange2021survey,momeni2025upper}.

Current continual learning methods can be broadly divided into
two families.
\emph{Rehearsal-based} methods keep a small buffer of past
examples and use it during new-task training --- replayed
directly~\citep{chaudhry2019er,rebuffi2017icarl}, with logit
anchoring~\citep{buzzega2020derpp}, selected for
retention~\citep{aljundi2019mir,aljundi2019gss}, or as gradient
constraints~\citep{chaudhry2019agem}.
\emph{Rehearsal-free} methods keep no data: they regularise
parameters toward past
solutions~\citep{kirkpatrick2017ewc,li2017lwf} or isolate tasks
in separate parameter subsets --- which in the LLM era usually
means separate LoRA adapters~\citep{hu2021lora}, kept
orthogonal~\citep{wang2023olora}, gated~\citep{liang2025gainlora},
or organised hierarchically~\citep{qian2025treelora}.  The
families differ in machinery but agree on the objective: both
answer the stability question, \emph{how do we keep the past
from being lost?}

\begin{figure}[t]
\centering
\resizebox{\linewidth}{!}{%
\begin{tikzpicture}[
  font=\small,
  box/.style={rectangle, draw, rounded corners, align=center,
              minimum height=0.75cm, inner sep=5pt},
  root/.style={box, fill=blue!10, minimum width=3.6cm, font=\small\bfseries},
  branch/.style={box, fill=blue!6, minimum width=3.0cm},
  q/.style={box, fill=blue!6, minimum width=3.9cm},
  carrier/.style={box, minimum width=3.6cm},
  arrow/.style={-Stealth, thick}
]
\node[root] (ct) at (0.8, 0) {Continual Learning};
\node[branch] (transfer) at (-2.2, -1.5)
  {\textbf{Transfer}\\``let the past help the new''};
\node[branch] (stab) at (6.0, -1.5)
  {\textbf{Stability}\\``preserve the past''};
\draw[arrow] (ct) -- (transfer);
\draw[arrow] (ct) -- (stab);
\node[q, fill=orange!12] (where) at (-6.4, -3.2)
  {\textbf{Where is it possible?}\\\scriptsize\itshape
   audit headroom};
\node[q, fill=orange!12] (what) at (-2.2, -3.2)
  {\textbf{What carries it?}\\\scriptsize\itshape
   a persistent carrier};
\node[q, fill=orange!12] (whom) at (2.0, -3.2)
  {\textbf{Whom to draw on?}\\\scriptsize\itshape
   route to the right source};
\draw[arrow] (transfer) -- (where);
\draw[arrow] (transfer) -- (what);
\draw[arrow] (transfer) -- (whom);
\node[carrier, fill=gray!16, draw=gray!60, dashed, minimum width=2.7cm]
  (replay) at (5.4, -3.2)
  {Replay\\\scriptsize\itshape conventional role:\\
   \scriptsize\itshape protect old tasks (\S\ref{sec:related})};
\node[carrier, fill=green!12, draw=green!55!black, very thick,
      minimum width=2.7cm]
  (kd) at (8.5, -3.2)
  {\textbf{Distillation (KD)}\\\scriptsize\itshape
   the stability term};
\draw[arrow] (stab) -- (replay);
\draw[arrow] (stab) -- (kd);
\node[carrier, fill=gray!16, draw=black!55] (param) at (-4.3, -5.1)
  {Parameter reuse\\\scriptsize\itshape extensively
   explored (\S\ref{sec:related})};
\node[carrier, fill=green!12, draw=green!55!black, very thick]
  (data) at (-0.1, -5.1)
  {\textbf{Data}\\\scriptsize\itshape this work: select
   \emph{which past}\\\scriptsize\itshape \emph{data} the new
   task trains with};
\draw[arrow] (what) -- (param);
\draw[arrow, green!55!black, very thick] (what) -- (data);
\draw[-Stealth, very thick, green!55!black, dashed, rounded corners=6pt]
  (replay.south) |- (data.east)
  node[pos=0.62, below, font=\scriptsize\itshape]
  {this work: point replay at the \emph{new} task as well};
\node[box, fill=green!10, draw=green!55!black, very thick,
      minimum width=12.6cm] (tsr) at (-1.2, -6.9)
  {\textbf{\method{} (one instantiation)}\\
   \scriptsize audited headroom \;$\cdot$\; selected replay
   \;$\cdot$\; distillation for stability};
\draw[arrow, green!55!black, very thick] (where.south) to[out=-90, in=170] (tsr.west);
\draw[arrow, green!55!black, very thick] (data.south) -- (tsr.north);
\draw[arrow, green!55!black, very thick] (whom.south) to[out=-90, in=75]
  ([xshift=2.1cm]tsr.north);
\end{tikzpicture}}%
\caption{\textbf{A framework for continual transfer, and one
instantiation.}  Continual learning serves two objectives.  On
the transfer side we organise the design space around three
questions --- where transfer is possible (headroom), what
carries it (parameters vs.\ data), and whom to draw on (source
selection); prior transfer mechanisms occupy the parameter
branch (\S\ref{sec:related}), and this work takes the data
branch, turning transfer into a data-selection problem.  On the
stability side, replay's conventional role is protecting old
tasks; we point the same machinery at the \emph{new} task as
well (dashed arrow) and assign the stability term to
distillation (\S\ref{sec:method:kd}).  \method{} instantiates
all three answers (\S\ref{sec:method}).}
\label{fig:paradigm}
\vspace{-4pt}
\end{figure}
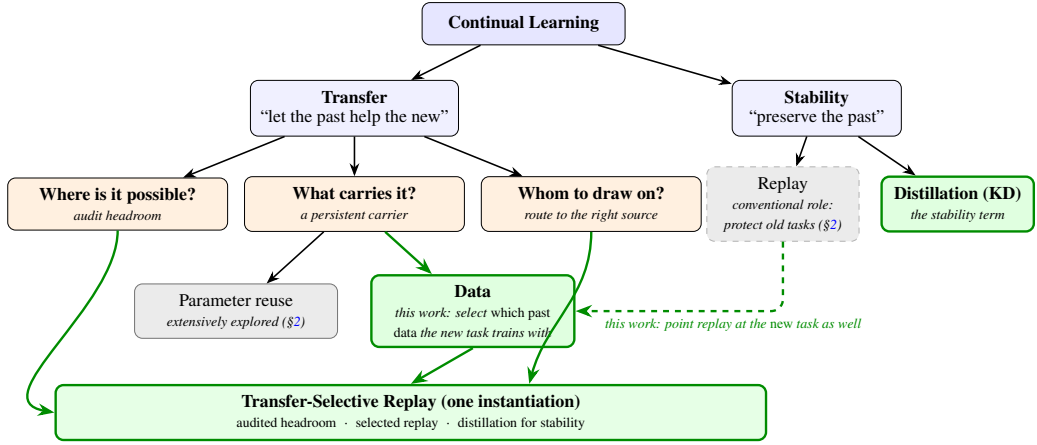

The complementary question --- whether past tasks can help
learning future ones --- has received far less attention.
\emph{Forward transfer}, the improvement in learning a new
task due to knowledge acquired from previous tasks, is the
promised upside of continual learning; it has been measured since
GEM~\citep{lopezpaz2017gem} and designed for since Progressive
Networks~\citep{rusu2016progressive}.  Yet on modern LLM
benchmarks it is routinely reported as absent or negative.
Existing attempts to promote forward transfer have relied
almost exclusively on \emph{parameter reuse}: lateral connections
and module reuse~\citep{rusu2016progressive,veniat2021ctrl},
task masks and shared adapters~\citep{ke2020cat,ke2021ctr},
prompt reuse~\citep{qin2022lfpt5}, and, for LLMs, routing,
initialising from, or merging past
adapters~\citep{huang2023lorahub,ilharco2022task,qian2025treelora}.
The assumption behind this line is so uniform that it is rarely
stated: \emph{past knowledge lives in past parameters}.
Past-task \emph{data}, meanwhile, has been used almost
exclusively to prevent forgetting rather than to improve
learning on new tasks.  Replay has been developed to preserve
old-task performance, while forward transfer has relied on
parameter reuse; the two directions have remained largely
separate.  And although negative transfer is increasingly
documented~\citep{merit2026,opr2026}, the field has no explicit
formulation of \emph{when} transfer should be expected at all.

This paper starts one step earlier: \emph{before designing a
transfer mechanism, we ask what makes transfer possible.}  We
study the question in \emph{low-budget} continual learning ---
each task arrives with a fixed, small training sample~\citep{zhao2024direct}.  
This is the continual-learning regime where forward transfer becomes
operationally valuable: with plentiful target data a task saturates on its own
and leaves nothing for the past to supply
(\S\ref{sec:paramfails}). Under a fixed budget, the value of
transfer is not to replace target supervision but to improve
the performance attainable from that budget by exploiting
previously acquired experience.

We identify three measurable conditions that determine whether
transfer is possible (\S\ref{sec:transfer}):
\begin{itemize}[leftmargin=1.4em, itemsep=1pt]
\item \emph{Where can transfer happen?}  Only where the target
  task leaves headroom beyond what its own limited supervision
  can achieve --- an auditable property before any transfer
  mechanism is introduced (\S\ref{sec:paramfails}).
\item \emph{What carries transfer?}  Only a persistent
  carrier.  Replay data repeatedly reinforces the same
  optimisation direction throughout training, whereas one-shot
  parameter reuse is progressively washed out
  (\S\ref{sec:paramfails:carrier}).
\item \emph{Whom should the new task draw on?}  Source quality
  varies sharply, but compatible sources can be identified at
  zero training cost using a label-aware \emph{task signature}
  (\S\ref{sec:whom}).
\end{itemize}
We organise the design space of continual transfer around these
three questions (Figure~\ref{fig:paradigm}).

\textbf{\method{}} (\methodshort{}) is one \emph{instantiation}
of this framework.
Choosing replay data as the carrier turns forward transfer into
a \emph{data-selection} problem: the central question is no
longer whether to replay, but which previous experience should
be replayed for the incoming task.
Concretely, \methodshort{} maintains a task memory holding one
record per past task --- a task signature, a small data
subsample, and the adapter snapshot from when that task ended.
When a new task arrives, a few forward--backward passes at a
fixed shared initialisation produce its signature at zero
training cost; a softmax over signature similarities then routes
every replay batch to the past data most aligned with the new
task, and each replayed batch is distilled against its own era's
snapshot.  \textbf{Distillation} complements replay by preserving
stability, allowing replay to focus on transfer --- a role our
analysis later identifies as optimisation regularisation rather
than knowledge preservation (\S\ref{sec:method:kd}).

We evaluate \methodshort{} under the standard
continual-learning protocol on TRACE-8~\citep{wang2023trace}
and NumGLUE-8~\citep{mishra2022numglue}, across three backbone
families (Qwen2.5~\citep{qwen25}, Gemma~3~\citep{gemma3}, and
Llama~3.2~\citep{llama3}) and thirteen state-of-the-art
continual-learning methods spanning both rehearsal-based and
rehearsal-free approaches.  \methodshort{} consistently
outperforms existing replay baselines by improving forward
transfer while maintaining positive backward transfer.

\paragraph{Contributions.}
\begin{enumerate}[leftmargin=1.6em]
\item \textbf{A framework for continual transfer.}  Three
  questions --- where (headroom), what (carrier), whom (source)
  --- each made measurable: benchmark headroom audits with
  oracle ceilings before mechanism design, a same-benchmark
  carrier dissociation, and the first pairwise data-transfer
  matrix in continual learning with LLMs, including its negative
  cells (\S\ref{sec:transfer}).
\item \textbf{An instantiation that redefines the role of replay.}
  The field's most reliable stability tool serves both
  objectives, and had so far been assigned only one:
  \methodshort{} points replay at the \emph{current} task,
  routed by a zero-training, label-aware signature over a task
  memory of (signature, data, snapshot) records.  Its selection matches the hindsight-oracle
  ceiling, and it is the strongest replay policy under the
  standard protocol (\S\ref{sec:method}--\ref{sec:experiments}).
\item \textbf{A clarification of the role of distillation in
  continual LLM learning.}  Through a systematic teacher audit
  --- identity, content, loss shaping, timing, snapshot source
  --- we show that distillation contributes through
  optimisation \emph{anchoring} rather than transferable
  teacher knowledge: its teachers matter as anchors, not as
  knowledge sources.  This dissociates the field's
  knowledge-preservation narrative from the operational
  mechanism, and
  motivates assigning distillation exclusively to the stability
  objective (\S\ref{sec:method:kd}, Appendix~\ref{app:kd}).
\end{enumerate}

\section{Related Work and Positioning}\label{sec:related}

\paragraph{Forward transfer.}  Transfer as an explicit goal of
continual learning is not new: \citet{lopezpaz2017gem}
formalised forward and backward transfer metrics, and designing
for forward transfer dates to Progressive
Networks~\citep{rusu2016progressive}.  GEM's forward transfer is
a zero-shot quantity (test before training); the training-time
sense studied here --- earlier tasks improving the
\emph{learning} of the current one --- is the promised upside of
sequential learning, and reusing source data for it is well
studied outside CL as intermediate-task training
(STILTs;~\citealp{phang2018stilts,vu2020exploring}) without ever
being instantiated inside a continual stream.  What is missing
is therefore not the objective but a \emph{formulation}: an
account of when transfer should exist, and how to measure it,
before a mechanism is designed (\S\ref{sec:transfer}).

\paragraph{Parameter-side transfer.}  The mechanisms proposed to realise
transfer during training are almost exclusively parameter-side:
module reuse~\citep{veniat2021ctrl}, masks and adapters over a
shared backbone
(CAT/B-CL/CTR;~\citealp{ke2020cat,ke2021bcl,ke2021ctr}),
meta-learned representations~\citep{javed2019oml}, and
soft-prompt reuse~\citep{qin2022lfpt5} --- a framing the NLP-CL
survey of \citet{keliu2022survey} makes explicit by listing all
transfer mechanisms as architectural.  In the LLM era the same
assumption drives inference-time composition (LoRAHub, task
arithmetic,
AdapterSoup;~\citealp{huang2023lorahub,ilharco2022task,
chronopoulou2023adaptersoup}) and continual-training
constraints: orthogonal adapters
(O-LoRA/N-LoRA;~\citealp{wang2023olora,zhao2025nlora}),
single-adapter merging~\citep{qiao2025slao}, and gated or routed
per-task
adapters~\citep{zhang2025clora,liang2025gainlora,ostapenko2024modular}.
These works answer the design question differently but share one
assumption: parameters are the transferable object.  This paper
investigates the complementary data-side formulation, and
\S\ref{sec:paramfails:carrier} measures where each carrier
survives --- inference-time composition is never trained
against, and nothing erases it.

\paragraph{Data replay.}  Replay is classically a
stability mechanism (ER, DER++,
iCaRL;~\citealp{chaudhry2019er,buzzega2020derpp,rebuffi2017icarl}),
and the replay-selection literature optimises the same
objective: MIR~\citep{aljundi2019mir} picks maximally interfered
samples, GSS~\citep{aljundi2019gss} gradient-diverse ones.  The
structurally closest works remain retention-directed:
MER~\citep{riemer2019mer} places a generic buffer inside a
transfer-maximising meta-objective without selecting \emph{which}
past task helps; Adaptive Memory
Replay~\citep{smith2024adaptivereplay} selects past-data clusters
conditioned on the current task, but its bandit reward is a
forgetting metric; InsCL~\citep{wang2024inscl} modulates replay
by task similarity in the \emph{inverse} direction (dissimilar
tasks get more replay, to protect them); and learned replay
controllers~\citep{proxymixing2026} frame replay as data
selection, again for retention.  Replay has been used to
preserve the past, not to accelerate the future --- the
objective this paper inverts (\S\ref{sec:method}).

Pointing replay at the new task then requires knowing which
past task will help --- a task-similarity question.  Fisher
embeddings
(Task2Vec;~\citealp{achille2019task2vec,vu2020exploring}),
optimal-transport distances
(OTDD;~\citealp{alvarezmelis2020otdd,sotdd2026}), gradient-based
task groupings~\citep{fifty2021tag}, and example-level gradient
selection (LESS;~\citealp{xia2024less}) all predict task
relationships; concurrent work predicts transfer from dataset
traits~\citep{krishna2025latent} or selects task
orderings~\citep{nguyen2025seqtrans}.  Closest inside CL,
TreeLoRA~\citep{qian2025treelora} routes new tasks to adapter
groups by gradient similarity at \emph{evolving} weights --- the
same class of signal, coupled to a parameter carrier; the
carrier dissociation of \S\ref{sec:paramfails:carrier} predicts
where that coupling holds.  The measured pairwise transfer
matrix of \S\ref{sec:whom} provides the prediction target these
representations have lacked: signatures are validated against
measured transfer rather than a routing heuristic.

\paragraph{Distillation.}  In continual learning, distillation
is narrated as knowledge preservation --- from LwF and
iCaRL~\citep{li2017lwf,rebuffi2017icarl} to DER++'s stored
logits~\citep{buzzega2020derpp} and recent
self-distillation~\citep{yang2024sdft}.  Adjacent fields already
read KL-anchoring as regularisation: reference-policy penalties
in RL~\citep{teh2017distral} and label-smoothing accounts of
supervised
distillation~\citep{yuan2020revisiting,mobahi2020self}.  Missing
is the continual-LLM dissociation --- whether the teacher's
\emph{knowledge}, or merely its existence as an \emph{anchor},
produces the gains; \S\ref{sec:method:kd} answers by audit.

\paragraph{Negative transfer.}  MERIT~\citep{merit2026} detects
gradient-level conflict in \emph{parallel} multi-task tuning;
OPR~\citep{opr2026} selects replayed rollouts by quality rather
than source-task identity; surveys note that replay can induce
negative forward transfer in fragile
domains~\citep{wang2024survey}.  These works document the
phenomenon.  Rather than documenting it, this paper predicts and
avoids it: the negative cells of the transfer matrix are exactly
what source selection routes around (\S\ref{sec:whom}).

\section{Problem Formulation}\label{sec:setup}

\subsection{Continual learning}

We consider a task-incremental continual learning setting: a learner faces a stream of tasks
$\mathcal{T} = (T_1, \dots, T_K)$, arriving one at a time, where
each task $T_t$ is a distribution $D_t$ over input--output pairs
and provides a training and a test sample from it.  We study
continual fine-tuning with a single accumulating parameter
state: the learner maintains one set of trainable parameters
$\theta$ throughout the stream; no per-task parameters are added
and no task identity (i.e., which task a test example belongs to) is available at inference.  When task $T_t$
arrives, sequential learning solves
\begin{equation}
\theta_t \;=\; \arg\min_{\theta}\;
\mathbb{E}_{(x,y)\sim D_t}\!\left[\,
\ell\!\left(f_\theta(x),\, y\right)\right],
\label{eq:cl}
\end{equation}
optimised from the previous state $\theta_{t-1}$; earlier tasks
are accessible only through bounded experience retained from the
stream (possibly none)~\citep{delange2021survey}.  Solving
Eq.~\eqref{eq:cl} naively is known to degrade earlier tasks ---
catastrophic forgetting --- and, less discussed, to forgo any
help earlier tasks could give the current one; continual-learning
methods differ in how they approximate Eq.~\eqref{eq:cl} under
this constraint (\S\ref{sec:method}).  Throughout, we study the
\emph{low-budget} regime: each incoming task provides only a
limited number of labeled examples, while experience from
earlier tasks is already in hand.  The concrete budget and the instantiation of the
parameter state are described in
\S\ref{sec:experiments:setup}.

\subsection{Transfer}\label{sec:setup:transfer}

Throughout this paper, \emph{forward transfer} refers to the
contribution of previously acquired experience to the learning
of the current task, under the fixed target-data budget.
Conceptually, we decompose current-task learning into
two contributions:
\begin{equation*}
\text{current-task learning} \;=\;
\underbrace{\text{target-data contribution}}_{\text{what the
task teaches itself}} \;+\;
\underbrace{\text{transfer contribution}}_{\text{what the past
supplies}}.
\end{equation*}
The transfer contribution is the object of this paper.  It is distinct
from GEM's zero-shot forward-transfer metric (test before
training)~\citep{lopezpaz2017gem} and from generalisation to
unseen tasks; and it is \emph{latent} --- no standard
continual-learning metric reads it off directly.  Since the contribution is not
directly observable, the remainder of the paper introduces
measurable surrogates: \S\ref{sec:transfer} constructs its
measurement and asks under what conditions it exists at all.

\section{Three Conditions for Realized Transfer}\label{sec:transfer}

We organize the investigation around a simple decomposition of
when forward transfer is realized in a continual stream:
\begin{equation}
\text{transfer}(j \to t) > 0
\;\;\text{requires}\;\;
\underbrace{\text{headroom}(t)}_{\textit{Where?}}
\;\wedge\;
\underbrace{\text{a persistent carrier}}_{\textit{What?}}
\;\wedge\;
\underbrace{\text{the right source } j}_{\textit{Whom?}},
\label{eq:transfer}
\end{equation}
We examine each condition in turn;
\S\ref{sec:method} then derives the method from them.

\subsection{Where: headroom as transfer opportunity}
\label{sec:paramfails}

Before asking how transfer can be realised, we must first ask
whether any transfer opportunity exists at all.

\begin{condition}\label{prop:headroom}
Without benchmark headroom, transfer is unmeasurable by any
mechanism: realised transfer is bounded by what additional
in-domain data buys.
\end{condition}

\paragraph{Headroom.}\label{sec:setup:measure}
The transfer contribution of \S\ref{sec:setup} is latent ---
training on mixed data yields one outcome in which target
learning and transfer are entangled --- so we first need a
diagnostic for \emph{where} it could appear at all.  Three
training runs under identical budgets --- the \emph{probe}
protocol used throughout --- define every quantity in the
paper: \textbf{target-data} (the $N$ target examples alone),
\textbf{target-full} ($10N$ target examples), and
\textbf{mixed}$(j)$ (a $50/50$ mixture of source task $j$'s
data and the $N$ target examples).  Writing $S_t(N)$ for
end-of-training test accuracy after training on $N$ of task $t$'s
own examples,
\begin{equation}
\mathrm{headroom}_t(N) \;=\; S_t(N_{\max}) \;-\; S_t(N),
\label{eq:headroom}
\end{equation}
with $N_{\max}$ a fixed well-resourced reference (ten times the
standard budget in our protocol): the gain that additional
\emph{in-domain} supervision --- the most valuable supervision
a task can receive --- would still buy at budget $N$.

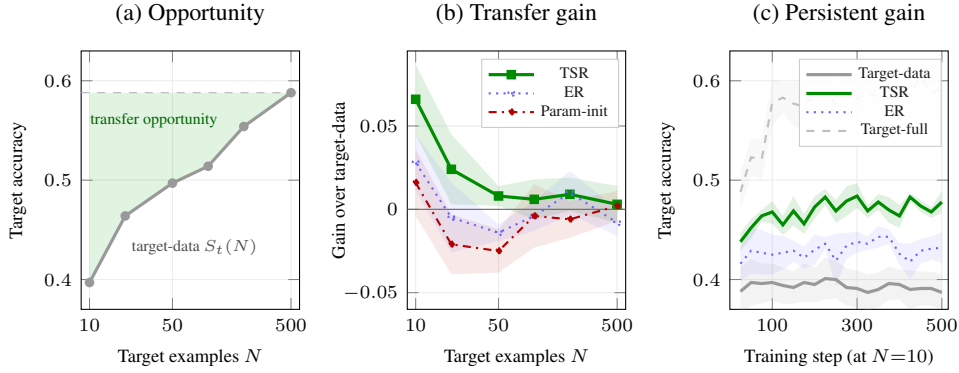
\begin{figure}[t]
\centering
\begin{tikzpicture}
\begin{axis}[
  name=left,
  width=0.32\linewidth, height=5.0cm,
  xmode=log, log ticks with fixed point,
  xtick={10,50,500},
  xlabel={Target examples $N$},
  ylabel={Target accuracy},
  xmin=8.5, xmax=590, ymin=0.37, ymax=0.63,
  title={\small (a) Opportunity},
  tick label style={font=\scriptsize},
  label style={font=\scriptsize},
  grid=major, grid style={gray!15},
]
\addplot[name path=A, gray!80, very thick, mark=*, mark size=1.2pt]
coordinates
{(10,.397) (20,.464) (50,.497) (100,.514) (200,.554) (500,.588)};
\addplot[name path=TOP, draw=none] coordinates
{(10,.588) (500,.588)};
\addplot[green!60!black, fill opacity=0.12, fill=green!60!black]
  fill between[of=A and TOP];
\node[font=\tiny, green!40!black] at (axis cs:35,.56)
  {transfer opportunity};
\node[font=\tiny, gray!60!black, anchor=north west]
  at (axis cs:19,.45) {target-data $S_t(N)$};
\addplot[gray!60, dashed, thin] coordinates {(8.5,.588) (590,.588)};
\end{axis}
\begin{axis}[
  at={(left.outer east)}, anchor=outer west, xshift=1mm, name=mid,
  width=0.32\linewidth, height=5.0cm,
  xmode=log, log ticks with fixed point,
  xtick={10,50,500},
  xlabel={Target examples $N$},
  ylabel={Gain over target-data},
  ylabel shift=-9pt,
  xmin=8.5, xmax=590, ymin=-0.06, ymax=0.095,
  title={\small (b) Transfer gain},
  legend style={font=\tiny, at={(0.97,0.97)}, anchor=north east,
                draw=gray!40, fill=white, fill opacity=0.9,
                inner xsep=2pt, inner ysep=1pt, row sep=-2.5pt},
  tick label style={font=\scriptsize},
  label style={font=\scriptsize},
  scaled y ticks=false,
  yticklabel style={/pgf/number format/fixed},
  grid=major, grid style={gray!15},
]
\addplot[black!60, thin, forget plot] coordinates {(8.5,0) (590,0)};
\addplot[name path=GU, draw=none, forget plot] coordinates
{(10,.087) (20,.045) (50,.014) (100,.018) (200,.019) (500,.014)};
\addplot[name path=GL, draw=none, forget plot] coordinates
{(10,.045) (20,.003) (50,.002) (100,-.006) (200,-.001) (500,-.008)};
\addplot[green!55!black, fill opacity=0.12, forget plot]
  fill between[of=GU and GL];
\addplot[name path=RU, draw=none, forget plot] coordinates
{(10,.045) (20,.016) (50,-.009) (100,.005) (200,.023) (500,.000)};
\addplot[name path=RL, draw=none, forget plot] coordinates
{(10,.011) (20,-.026) (50,-.019) (100,-.013) (200,-.003) (500,-.016)};
\addplot[blue!60, fill opacity=0.10, forget plot]
  fill between[of=RU and RL];
\addplot[name path=PU, draw=none, forget plot] coordinates
{(10,.036) (20,-.003) (50,-.012) (100,.015) (200,.005) (500,.011)};
\addplot[name path=PL, draw=none, forget plot] coordinates
{(10,-.004) (20,-.039) (50,-.038) (100,-.023) (200,-.017) (500,-.007)};
\addplot[red!70!black, fill opacity=0.09, forget plot]
  fill between[of=PU and PL];
\addplot[green!55!black, very thick, mark=square*, mark size=1.2pt]
coordinates
{(10,.066) (20,.024) (50,.008) (100,.006) (200,.009) (500,.003)};
\addlegendentry{TSR}
\addplot[blue!60, thick, dotted, mark=o, mark size=1.1pt] coordinates
{(10,.028) (20,-.005) (50,-.014) (100,-.004) (200,.010) (500,-.008)};
\addlegendentry{ER}
\addplot[red!70!black, thick, dash dot, mark=diamond*, mark size=1.3pt]
coordinates
{(10,.016) (20,-.021) (50,-.025) (100,-.004) (200,-.006) (500,.002)};
\addlegendentry{Param-init}
\end{axis}
\begin{axis}[
  at={(mid.outer east)}, anchor=outer west, xshift=1mm,
  width=0.32\linewidth, height=5.0cm,
  xlabel={Training step (at $N{=}10$)},
  ylabel={Target accuracy},
  xtick={100,300,500},
  legend style={font=\tiny, at={(0.97,0.97)}, anchor=north east,
                draw=gray!40, fill=white, fill opacity=0.9,
                inner xsep=2pt, inner ysep=1pt, row sep=-2.5pt},
  xmin=0, xmax=515, ymin=0.37, ymax=0.63,
  title={\small (c) Persistent gain},
  tick label style={font=\scriptsize},
  label style={font=\scriptsize},
  grid=major, grid style={gray!15},
]
\addplot[name path=CAU, draw=none, forget plot] coordinates
{(25,0.412) (50,0.421) (75,0.416) (100,0.416) (125,0.412) (150,0.408) (175,0.416) (200,0.412) (225,0.422) (250,0.419) (275,0.411) (300,0.409) (325,0.405) (350,0.407) (375,0.413) (400,0.410) (425,0.409) (450,0.408) (475,0.408) (500,0.405)};
\addplot[name path=CAL, draw=none, forget plot] coordinates
{(25,0.364) (50,0.374) (75,0.376) (100,0.379) (125,0.375) (150,0.376) (175,0.378) (200,0.377) (225,0.380) (250,0.381) (275,0.374) (300,0.372) (325,0.369) (350,0.372) (375,0.379) (400,0.380) (425,0.370) (450,0.374) (475,0.374) (500,0.369)};
\addplot[gray!70, fill opacity=0.13, forget plot] fill between[of=CAU and CAL];
\addplot[name path=CGU, draw=none, forget plot] coordinates
{(25,0.447) (50,0.461) (75,0.470) (100,0.479) (125,0.464) (150,0.484) (175,0.466) (200,0.484) (225,0.491) (250,0.480) (275,0.494) (300,0.497) (325,0.475) (350,0.486) (375,0.483) (400,0.475) (425,0.487) (450,0.480) (475,0.473) (500,0.489)};
\addplot[name path=CGL, draw=none, forget plot] coordinates
{(25,0.429) (50,0.442) (75,0.458) (100,0.456) (125,0.445) (150,0.454) (175,0.446) (200,0.461) (225,0.475) (250,0.457) (275,0.464) (300,0.472) (325,0.462) (350,0.470) (375,0.457) (400,0.452) (425,0.479) (450,0.468) (475,0.462) (500,0.467)};
\addplot[green!55!black, fill opacity=0.12, forget plot] fill between[of=CGU and CGL];
\addplot[name path=CRU, draw=none, forget plot] coordinates
{(25,0.433) (50,0.452) (75,0.450) (100,0.445) (125,0.444) (150,0.439) (175,0.433) (200,0.445) (225,0.443) (250,0.446) (275,0.447) (300,0.448) (325,0.443) (350,0.454) (375,0.450) (400,0.437) (425,0.436) (450,0.441) (475,0.443) (500,0.448)};
\addplot[name path=CRL, draw=none, forget plot] coordinates
{(25,0.399) (50,0.406) (75,0.405) (100,0.405) (125,0.410) (150,0.418) (175,0.411) (200,0.415) (225,0.430) (250,0.393) (275,0.421) (300,0.427) (325,0.430) (350,0.432) (375,0.436) (400,0.415) (425,0.400) (450,0.416) (475,0.420) (500,0.417)};
\addplot[blue!60, fill opacity=0.10, forget plot] fill between[of=CRU and CRL];
\addplot[name path=CDU, draw=none, forget plot] coordinates
{(25,0.507) (50,0.542) (75,0.541) (100,0.602) (125,0.598) (150,0.599) (175,0.597) (200,0.610) (225,0.609) (250,0.594) (275,0.607) (300,0.602) (325,0.603) (350,0.607) (375,0.607) (400,0.608) (425,0.610) (450,0.624) (475,0.616) (500,0.619)};
\addplot[name path=CDL, draw=none, forget plot] coordinates
{(25,0.469) (50,0.505) (75,0.501) (100,0.554) (125,0.568) (150,0.554) (175,0.551) (200,0.564) (225,0.568) (250,0.560) (275,0.567) (300,0.562) (325,0.574) (350,0.562) (375,0.559) (400,0.561) (425,0.564) (450,0.582) (475,0.561) (500,0.565)};
\addplot[gray!50, fill opacity=0.10, forget plot] fill between[of=CDU and CDL];
\addplot[gray!80, very thick] coordinates {(25,0.388) (50,0.397) (75,0.396) (100,0.397) (125,0.394) (150,0.392) (175,0.397) (200,0.395) (225,0.401) (250,0.4) (275,0.392) (300,0.391) (325,0.387) (350,0.39) (375,0.396) (400,0.395) (425,0.39) (450,0.391) (475,0.391) (500,0.387)};
\addlegendentry{Target-data}
\addplot[green!55!black, very thick] coordinates {(25,0.438) (50,0.452) (75,0.464) (100,0.468) (125,0.455) (150,0.469) (175,0.456) (200,0.473) (225,0.483) (250,0.469) (275,0.479) (300,0.484) (325,0.469) (350,0.478) (375,0.47) (400,0.464) (425,0.483) (450,0.474) (475,0.468) (500,0.478)};
\addlegendentry{TSR}
\addplot[blue!60, thick, dotted] coordinates {(25,0.416) (50,0.429) (75,0.427) (100,0.425) (125,0.427) (150,0.429) (175,0.422) (200,0.43) (225,0.436) (250,0.419) (275,0.434) (300,0.438) (325,0.436) (350,0.443) (375,0.443) (400,0.426) (425,0.418) (450,0.429) (475,0.431) (500,0.432)};
\addlegendentry{ER}
\addplot[gray!50, thick, dashed] coordinates {(25,0.488) (50,0.523) (75,0.521) (100,0.578) (125,0.583) (150,0.577) (175,0.574) (200,0.587) (225,0.588) (250,0.577) (275,0.587) (300,0.582) (325,0.588) (350,0.584) (375,0.583) (400,0.584) (425,0.587) (450,0.603) (475,0.588) (500,0.592)};
\addlegendentry{Target-full}
\end{axis}
\end{tikzpicture}
\caption{\textbf{Transfer opportunity shrinks with the target
budget, and the gain lives where opportunity is large}
(flagship target numglue-cm, the matrix's strongest pair;
$n{=}5$ seeds per point;
shading $\pm 1$ s.e.m.).  \emph{(a)} headroom vs.\ budget;
\emph{(b)} paired transfer gain over target-data (TSR denotes
selected replay, ER uniform replay); \emph{(c)} at
$N{=}10$ the gain holds from the first evaluation to the last
step.  Per-target numbers: Appendix~\ref{app:jointtablesec}.}
\label{fig:budget}
\end{figure}

\paragraph{Interpretation.}
Headroom measures how much improvement remains unavailable from
the task's own budget --- \emph{task saturation}, and through
it the \emph{transfer opportunity}.
Figure~\ref{fig:budget}(a) is the evidence: the target-data
curve rises from $.397$ at $N{=}10$ to $.588$ at $N{=}500$, so
headroom is not a fixed property of a benchmark but a function
of the target-data budget --- large exactly where labels are
scarce.

\paragraph{Quantification.}
Since the contribution is latent, we report its practical
effect as \emph{equivalent target supervision} (a
sample-efficiency reading): a measured gain at budget $N$ is
converted, through the $S_t(N)$ curve, into the number of
additional target labels that would buy the same improvement.
This is the long-standing convention in transfer measurement
--- samples-to-threshold in RL
transfer~\citep{taylor2009transfer}, source ranking under
limited target supervision in
taskonomy~\citep{zamir2018taskonomy}, and \emph{effective data
transferred} in neural scaling
laws~\citep{hernandez2021scaling}.
Figure~\ref{fig:budget}(b): at $N{=}10$, selected mixing gains
$+.066$ --- the performance of ${\approx}20$ target examples, a
$2\times$ label efficiency, $35\%$ of the remaining opportunity
--- and the gain decays as saturation closes, never crossing
zero on this target; across all five classification targets the
same pattern holds with smaller amplitude, turning negative
once headroom is spent.  Panel (c) shows the gain is present
from the first evaluation and sustained for all $500$ steps ---
not an endpoint artefact (per-target numbers are in
Appendix~\ref{app:jointtablesec}).  The two other arms preview the
remaining conditions: uniform replay (ER) captures less than
half the low-budget gain and sits below selection at every budget
(\S\ref{sec:whom}); the parameter-init arm never separates from
zero (\S\ref{sec:paramfails:carrier}).

\paragraph{Implication.}
A benchmark with zero headroom cannot reveal transfer, whatever
the mechanism.  On the canonical 15-task suite of
\citet{razdaibiedina2023prog} (PP15;
\citealp{wang2023olora,qin2022lfpt5}), an instruction-tuned
0.5B base leaves essentially none: ten times more target data
buys $+0.02$ on average, and an audit of fourteen
parameter-reuse configurations with an oracle-selected source
lands within one seed standard deviation of the plain baseline
(Appendix~\ref{app:pp15}).  TRACE-8~\citep{wang2023trace}, by
contrast, retains real headroom at the same scale --- hence our
benchmark choice (\S\ref{sec:whom} draws the benchmark-design
corollary).
\citet{wang2023trace} argued qualitatively that classic
suites lack challenge for aligned LLMs, and
\citet{plateau2026} recommend reporting saturation statistics;
the headroom audit is the actionable, budget-conditioned
version.  Our merge-protocol artefact --- an apparent transfer
signal $35\times$ its protocol-clean value --- joins the
documented genus of evaluation
artefacts~\citep{rethinkmerge2024,zheng2025spurious,pseudoforget2024}.

\subsection{What: the carrier must persist}
\label{sec:paramfails:carrier}

Headroom tells us whether transfer is \emph{possible}; it does
not tell us how the transferred signal reaches the learner.
This motivates the next question: \emph{what carries transfer?}
A carrier can take many forms --- parameters, data, prompts,
external memory --- and the stream constrains it only to be
something retained from past tasks.  Whatever its form, the carrier's signal must survive the
continual optimisation.

\begin{condition}\label{prop:carrier}
Realised transfer requires a carrier that preserves its signal
throughout optimisation; a transient carrier is erased
regardless of headroom or source quality.
\end{condition}

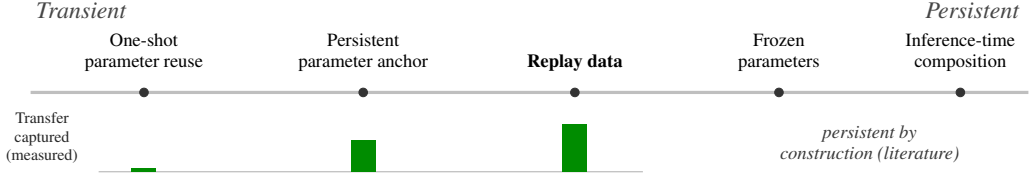
\begin{figure}[t]
\centering
\begin{tikzpicture}[x=1cm]
\draw[very thick, gray!50] (0,0) -- (13.2,0);
\node[font=\small\itshape, gray!55!black, anchor=south west]
  at (-0.05,0.86) {Transient};
\node[font=\small\itshape, gray!55!black, anchor=south east]
  at (13.2,0.86) {Persistent};
\foreach \x in {1.5,4.4,7.2,9.9,12.3}
  \filldraw[gray!40!black] (\x,0) circle (1.6pt);
\node[font=\scriptsize, align=center, anchor=south] at (1.5,0.16)
  {One-shot\\parameter reuse};
\node[font=\scriptsize, align=center, anchor=south] at (4.4,0.16)
  {Persistent\\parameter anchor};
\node[font=\scriptsize, align=center, anchor=south] at (7.2,0.16)
  {\textbf{Replay data}};
\node[font=\scriptsize, align=center, anchor=south] at (9.9,0.16)
  {Frozen\\parameters};
\node[font=\scriptsize, align=center, anchor=south] at (12.3,0.16)
  {Inference-time\\composition};
\draw[gray!60] (0.9,-1.05) -- (8.1,-1.05);
\fill[green!55!black] (1.34,-1.05) rectangle (1.66,-1.01);
\fill[green!55!black] (4.24,-1.05) rectangle (4.56,-0.63);
\fill[green!55!black] (7.04,-1.05) rectangle (7.36,-0.42);
\node[font=\tiny, align=center, anchor=east, gray!30!black]
  at (0.75,-0.60) {Transfer\\captured\\(measured)};
\node[font=\scriptsize\itshape, gray!45!black, align=center]
  at (11.1,-0.68) {persistent by\\construction (literature)};
\end{tikzpicture}
\caption{\textbf{Candidate carriers, ordered by the persistence
of their optimization signal.}  The transfer gain follows
persistence rather than whether the carrier is parameters or
data; bars show the measured share of the data-mixing accuracy gain each carrier captures (Appendix~\ref{app:ablations}).  The
literature examples shown here are persistent by
construction.}
\label{fig:carrieraxis}
\end{figure}

Figure~\ref{fig:carrieraxis} tests the requirement across
candidate carriers on TRACE-8: identical budgets, the same
source, different carriers.  A
transient carrier --- one-shot initialisation from the
source-trained adapter --- captures none of the mixing gain;
its early advantage is real but erased by continued training
(\emph{wash-out}).  A
persistent \emph{parameter} carrier --- an L2 anchor toward the
same source adapter, kept in the loss at every step ---
captures roughly two-thirds (Appendix~\ref{app:ablations}).
Persistent \emph{data} --- the source's examples mixed into
every batch --- captures all of it.  The transfer gain thus
rises with persistence, not with whether the carrier is
parameters or data.  The same axis organises the literature:
frozen columns~\citep{rusu2016progressive} and inference-time
composition~\citep{huang2023lorahub} succeed and are persistent
by construction --- the source signal is never optimised away
--- whereas merge-then-train pipelines re-enter at the
transient end; the literature appears contradictory only
because it mixes transient and persistent parameter mechanisms
(Appendix~\ref{app:scope}).

Among the carriers we evaluate, replay data is the simplest
persistent one: no parameter growth, no architectural change,
no task identity at inference.  Data is therefore not
privileged because it is data, but because it is the simplest
carrier whose signal is reintroduced throughout optimisation
--- the premise from which \S\ref{sec:method} derives the
method, and which \S\ref{sec:experiments} stress-tests at
benchmark scale.

\subsection{Whom: source selection}\label{sec:whom}

Knowing that transfer is possible, and knowing what carries it,
still leaves one question open: which previous task should the
learner draw on?

\begin{condition}\label{prop:source}
Given headroom and a persistent carrier, realised transfer is
determined by source choice --- including its sign; and the
choice is predictable from label-aware learning directions at
the shared initialisation.
\end{condition}

\subsubsection{The transfer matrix}\label{sec:matrix}

We measure the full transfer matrix over all ordered pairs of
TRACE-8 (Table~\ref{tab:matrix}).

\begin{table}[t]
\centering\small
\begin{tabular}{l|cccccccc}
\toprule
source$\backslash$target & cstn & fomc & meet & py15 & sciq & ngcm & ngds & 20mn \\
\midrule
cstance     & --- & \cellcolor{green!15}$\mathbf{-0.18}$ & \cellcolor{green!15}$-0.20$ & \cellcolor{green!26}$-0.70$ & \cellcolor{red!26}$+0.19$ & \cellcolor{gray!6}$-0.01$ & \cellcolor{green!26}$-0.69$ & \cellcolor{green!26}$-0.63$ \\
fomc        & \cellcolor{green!7}$\mathbf{-0.07}$ & --- & \cellcolor{green!15}$-0.24$ & \cellcolor{green!26}$-0.45$ & \cellcolor{green!7}$\mathbf{-0.09}$ & \cellcolor{red!13}$+0.14$ & \cellcolor{green!15}$-0.27$ & \cellcolor{green!26}$-0.69$ \\
meetingbank & \cellcolor{gray!6}$+0.02$ & \cellcolor{gray!6}$-0.02$ & --- & \cellcolor{green!15}$-0.34$ & \cellcolor{red!13}$+0.12$ & \cellcolor{green!15}$\mathbf{-0.26}$ & \cellcolor{green!26}$-0.53$ & \cellcolor{green!26}$-0.45$ \\
py150       & \cellcolor{gray!6}$-0.03$ & \cellcolor{green!15}$-0.15$ & \cellcolor{green!15}$-0.29$ & --- & \cellcolor{red!26}$+0.16$ & \cellcolor{green!7}$-0.10$ & \cellcolor{green!15}$-0.28$ & \cellcolor{green!26}$-0.46$ \\
scienceqa   & \cellcolor{gray!6}$-0.04$ & \cellcolor{green!7}$-0.09$ & \cellcolor{green!15}$\mathbf{-0.36}$ & \cellcolor{green!26}$\mathbf{-0.76}$ & --- & \cellcolor{red!26}$+0.26$ & \cellcolor{green!26}$-0.47$ & \cellcolor{green!26}$-0.75$ \\
numglue-cm  & \cellcolor{gray!6}$+0.01$ & \cellcolor{gray!6}$-0.04$ & \cellcolor{green!15}$-0.21$ & \cellcolor{green!26}$-0.68$ & \cellcolor{gray!6}$-0.03$ & --- & \cellcolor{green!26}$\mathbf{-0.88}$ & \cellcolor{green!26}$\mathbf{-0.85}$ \\
numglue-ds  & \cellcolor{gray!6}$+0.01$ & \cellcolor{green!7}$-0.11$ & \cellcolor{green!15}$-0.29$ & \cellcolor{green!26}$-0.62$ & \cellcolor{gray!6}$-0.01$ & \cellcolor{green!15}$-0.20$ & --- & \cellcolor{green!26}$-0.80$ \\
20minuten   & \cellcolor{red!13}$+0.06$ & \cellcolor{gray!6}$-0.02$ & \cellcolor{green!15}$-0.18$ & \cellcolor{green!15}$-0.35$ & \cellcolor{red!26}$+0.27$ & \cellcolor{green!15}$-0.16$ & \cellcolor{gray!6}$+0.00$ & --- \\
\bottomrule
\end{tabular}
\caption{Transfer matrix on TRACE-8 (pairwise joint
training of source $j$ mixed $50/50$ with $N{=}50$ target
examples vs.\ target-data; mean of $5$ seeds; bold = column
best).  \textcolor{green!55!black}{Green} = source helps the
target (darker = larger loss reduction);
\textcolor{red!70!black}{red} = source \emph{hurts}.}
\label{tab:matrix}
\end{table}

\textbf{Three structural findings.}
(1)~\emph{Semantic pairs are mutual best sources}:
fomc $\leftrightarrow$ cstance are each other's best source;
NumGLUE-cm is NumGLUE-ds's best at $-0.88$, the largest cell.
(2)~\emph{Negative transfer is real and target-specific}:
cstance, scienceqa, and numglue-cm are fragile targets where the
wrong source actively hurts (worst cells $+0.27$, $+0.26$) ---
the empirical basis for \S\ref{sec:method}'s headline.
(3)~\emph{Generation targets absorb anything} ($-0.2$ to
$-0.85$ from nearly all sources): pair-specific selection is a
classification-task phenomenon here.

\paragraph{Benchmark-design corollary.}  Contrasting the two
benchmarks: \emph{a CL benchmark should report its data headroom
(at a stated data budget) and transfer structure as part of
its specification}.  PP15-class suites evaluated with
instruction-tuned bases have neither, and any transfer comparison
run on them inherits a ceiling of zero regardless of method
(Appendix~\ref{app:pp15}) --- which may explain why routing and
reuse comparisons in this
literature so often report method-insensitive results.

\subsubsection{Task signatures}\label{sec:transfer:predict}

The matrix defines what a selector must predict; measuring it
requires training every ordered source--target pair ---
feasible for analysis, impractical in continual learning, where
the source must be chosen the moment a new task arrives.  We
therefore seek a \emph{task signature}: a training-free
representation of each task from which measured transfer can be
predicted.  The candidates we evaluate span three increasingly
informative families --- input similarity, dataset geometry,
and learning direction (Table~\ref{tab:predictors}).  Only the
\emph{gradient signature} consistently identifies the best
source:
\begin{equation}
g_t \;=\; \frac{1}{m} \sum_{b=1}^{m}
\nabla_{B}\, \mathcal{L}_\mathrm{CE}(b)
\Big\rvert_{(A_0, B_0)},
\qquad
\mathrm{score}(j, t) = \cos(g_j, g_t),
\label{eq:signature}
\end{equation}
Here $g_t$ is the mean gradient over the $m$ probe batches,
taken at the fixed shared initialisation $(A_0, B_0)$ and
flattened across layers; $B$ is the zero-initialised LoRA
factor, so $g_t$ is the adapter's entire first-order learning
direction, and sources are scored by the cosine between
directions.

\paragraph{Mechanism.}  The signature is not a semantic
representation: it captures the \emph{optimisation bias} a
task induces at the shared initialisation --- the direction in
which the task asks the adapter to change --- so it reflects
what a task asks the model to \emph{learn}, not what its data
looks like.  This is also why alignment should predict
transfer under replay.  Replayed batches act on the same
accumulating adapter as the target's own batches: gradient
steps from an aligned source reinforce the trajectory the
target demands --- the replay supplies optimisation bias
rather than additional supervision --- whereas steps from a
conflicting source compete with the target update for the same
parameters.  If this interpretation is correct, aligned source
updates should produce more compatible descent on the target
than conflicting ones.

\paragraph{Local intervention.}  We test this optimisation
hypothesis by intervention.  Holding the adapter at the shared
initialisation --- the state where signatures are taken --- we
apply one target update, add an equal-norm gradient
perturbation from a past task --- so that only the replay
direction, not its magnitude, differs across sources --- and
measure the loss change on an independent held-out target
batch, isolating its effect from additional target supervision
or long-horizon optimisation (protocol in
Appendix~\ref{app:intervention}).
Sources grouped by the stored signature produce strictly
ordered marginal descent (Table~\ref{tab:predictors}, right),
demonstrating that replay directions selected by the signature
provide progressively more compatible optimisation steps.
Current-gradient alignment predicts the immediate reduction
within target (Spearman $\rho = .87$), and the stored
signature preserves this geometry ($\rho = .85$ against
current-gradient cosines) --- routing on the stored signature
therefore predicts compatible descent rather than input
similarity;
\S\ref{sec:experiments} tests whether this local effect
survives repeated optimisation.

\begin{table}[t]
\centering\small
\begin{tabular}{llc}
\multicolumn{3}{c}{\emph{Prediction}} \\
\toprule
Signature & Signal & Top-1 \\
\midrule
Proto        & input similarity            & $1/8$ \\
HiddenOT     & distribution geometry       & $2/8$ \\
JointOT      & label-aware distribution    & $0/8$ \\
Fisher       & gradient magnitude          & $1/8$ \\
GradB\_last4 & gradient direction (last $4$ blocks) & $3/8$ \\
GradB\_norm  & gradient direction (normalised)      & $\mathbf{4/8}$ \\
\textbf{GradB} & gradient direction (Eq.~\ref{eq:signature}) & $\mathbf{4/8}$ \\
\bottomrule
\end{tabular}\hspace{14pt}%
\begin{tabular}{lc}
\multicolumn{2}{c}{\emph{Mechanism}} \\
\toprule
Added update & marginal $\Delta L$ \\
\midrule
Extra target batch & $-1.08$ \\
Top-aligned source & $-.30$ \\
Middle source      & $-.12$ \\
Bottom source      & $-.02$ \\
\bottomrule
\end{tabular}
\caption{\textbf{Task signatures evaluated against the measured transfer matrix of Table~\ref{tab:matrix}.}
Deployment selects one source, so we score by best-source top-1
hit rate; robustness checks are in
Appendix~\ref{app:signature}.  JointOT is our OTDD-style
label-aware distributional baseline~\citep{alvarezmelis2020otdd,
sotdd2026}.  Right: local intervention at the
shared initialisation (three classification targets; protocol
in Appendix~\ref{app:intervention}).}
\label{tab:predictors}
\end{table}

\paragraph{Takeaways.}
(i)~\textbf{The gradient signature is the only reliable argmax
predictor}: it identifies the oracle-best source for $4/8$
targets overall and $4/5$ classification targets; its one miss
selects the matrix's second-best source.
(ii)~\textbf{Direction, not magnitude, carries the signal}:
squaring gradients (Fisher/task2vec-style) destroys it;
normalising per-batch directions reproduces it exactly.
(iii)~\textbf{Functional beats distributional}:
input-similarity and distribution-geometry signatures find the
best source for at most $2/8$ targets, and adding labels does
not fix it (JointOT: top-1 $0/8$ despite ranking well) --- the failures
mirror the mechanism: distributional signatures ignore what
tasks ask the model to \emph{do}, and gradient magnitudes
ignore \emph{where} the demands point.  Further analyses are reported in
Appendix~\ref{app:signature}.

With measurable answers to \emph{where}, \emph{what}, and
\emph{whom}, we now construct one algorithmic instantiation.

\section{\method{}}\label{sec:method}

\methodshort{} is one instantiation of the \S\ref{sec:transfer}
framework.  The three conditions enter at two different levels:
headroom admits no algorithm --- it is a property of the
benchmark, base, and budget, settled by the audit before any
mechanism is designed (\S\ref{sec:paramfails}) --- so the
algorithm instantiates the remaining two, carrying transfer
with persistent data (\S\ref{sec:paramfails:carrier}) and
choosing sources by task signature (\S\ref{sec:whom}), while
distillation takes over the stability objective that replay no
longer serves (\S\ref{sec:method:kd}).  We present the general
form first (Algorithm~\ref{alg:tsr}), then the instantiation
evaluated in this paper (Algorithm~\ref{alg:tsrt},
Figure~\ref{fig:pipeline}).

\subsection{The general algorithm}\label{sec:method:alg}

\textbf{Algorithm~\ref{alg:tsr}} is what the framework
requires of \emph{any} instantiation: standard sequential training of an
adapter, plus three replaceable modules --- \textsc{Probe}
(represent the incoming task before training), \textsc{Route}
(weigh past tasks by predicted transfer), and \textsc{Reference}
(anchor each batch for stability) --- with transfer carried by
routed replay in every optimisation step.  The modules are
intentionally unspecified here; \S\ref{sec:method:inst} fixes
them.

\paragraph{Notation.}  Both algorithms share the following
symbols.  $D_t$ is task $t$'s training distribution, and $b$ a
minibatch; $\theta$ is the trainable adapter, $\theta_0$ the
shared initialisation, $\theta_t$ the snapshot frozen when task
$t$ ends.  Each past task $j$ leaves one record
$(g_j,\, \mathcal{M}_j,\, \theta_j)$ --- signature, stored data
subsample, snapshot --- in the task memory $\mathcal{P}$; $w$
is the routing distribution over these records, and
$\theta^{\mathrm{ref}}$ the frozen reference (teacher) of the
stability term.  $\mathcal{L}_\mathrm{CE}$ is the task loss of
Eq.~\eqref{eq:cl}, $\mathcal{L}_\mathrm{KD}$ the stability
regulariser (\S\ref{sec:method:kd}), $\nabla_B$ the gradient
w.r.t.\ the adapter's zero-initialised low-rank factor $B$
(Eq.~\ref{eq:signature}), and $f_\theta$ the model's
token-level predictive distribution.  Hyperparameters: replay
ratio $\alpha$, stability weight $\lambda$, routing temperature
$\tau$, probe batches $m$, learning rate $\eta$.

\begin{algorithm}[t]
\caption{General \methodshort{}}
\label{alg:tsr}
\begin{algorithmic}[1]
\Require task stream $T_1, \dots, T_K$ with training data
  $D_1, \dots, D_K$; initial parameters $\theta_0$; modules
  \textsc{Probe}, \textsc{Route}, \textsc{Reference};
  hyperparameters $\alpha$, $\lambda$, $\eta$
\State $\mathcal{P} \gets \{\}$;\quad $\theta \gets \theta_0$
\For{$t = 1, \dots, K$}
  \State \colorbox{yellow!25}{$g_t \gets \textsc{Probe}(D_t)$}
  \State \colorbox{yellow!25}{$w \gets \textsc{Route}\big(g_t,\
    \{g_j\}_{j<t}\big)$}
  \For{each training step}
    \State with prob.\ $1{-}\alpha$: $b \sim D_t$;\quad
      otherwise: $j \sim w$,\ $b \sim \mathcal{M}_j$
    \State \colorbox{yellow!25}{$\theta^{\mathrm{ref}} \gets
      \textsc{Reference}(b,\ \mathcal{P})$}
    \State $\theta \gets \theta - \eta\, \nabla_\theta
      \big[\, \mathcal{L}_\mathrm{CE}(\theta; b) + \lambda\,
      \mathcal{L}_\mathrm{KD}(\theta;\, \theta^{\mathrm{ref}},
      b) \,\big]$
  \EndFor
  \State $\mathcal{M}_t \gets$ subsample of $D_t$;\quad
    $\theta_t \gets \theta$;\quad
    $\mathcal{P} \gets \mathcal{P} \cup
    \{(g_t,\, \mathcal{M}_t,\, \theta_t)\}$
\EndFor
\end{algorithmic}
\end{algorithm}

The organising object is the \emph{task memory}
$\mathcal{P} = \{(g_j,\, \mathcal{M}_j,\, \theta_j)\}_{j<t}$:
one record per past task, written once when the task ends and
never updated.  Its three fields answer an incoming task's
three needs --- the signature says \emph{whether} $j$ should be
drawn on, the data subsample is \emph{what} gets replayed, and
the era snapshot is \emph{who} supervises that replay.  Because
the fields are bound in one record, a single routing decision
returns coherent (data, teacher) pairs: replayed knowledge
always arrives with the reference model that knew it best.

\subsection{The evaluated instantiation}\label{sec:method:inst}

\textbf{Algorithm~\ref{alg:tsrt}} --- the \emph{routed
triple} --- fixes the three modules; the evaluated
hyperparameters are listed with the experimental setup
(\S\ref{sec:experiments:setup}).  Figure~\ref{fig:pipeline}
traces one task through the resulting pipeline; each choice
carries its rationale from \S\ref{sec:transfer}.

\begin{algorithm}[t]
\caption{Routed-triple \methodshort{}}
\label{alg:tsrt}
\begin{algorithmic}[1]
\Require task stream $T_1, \dots, T_K$ with training data
  $D_1, \dots, D_K$; shared initialisation $\theta_0$;
  hyperparameters $m$, $\tau$, $\alpha$, $\lambda$, $\eta$
\State $\mathcal{P} \gets \{\}$;\quad $\theta \gets \theta_0$
\For{$t = 1, \dots, K$}
  \State \colorbox{yellow!25}{$g_t \gets \frac{1}{m}
    \sum_{i=1}^{m} \nabla_B\, \mathcal{L}_\mathrm{CE}(\theta_0;
    b_i)$, \ $b_i \sim D_t$}
    \Comment{\textsc{Probe}}
  \State \colorbox{yellow!25}{$w_j \gets
    \mathrm{softmax}_{j<t}\!\big(\cos(g_j,\, g_t) / \tau\big)$}
    \Comment{\textsc{Route}}
  \State $j^\ast \gets \arg\max_{j<t} w_j$
  \For{each training step}
    \State draw $u \sim \mathrm{Uniform}(0,1)$
    \If{$u > \alpha$}
      \State sample minibatch $b \sim D_t$;\quad
        \colorbox{yellow!25}{$\theta^{\mathrm{ref}} \gets
        \theta_{j^\ast}$}
        \Comment{current-task batch}
    \Else
      \State draw one source $j \sim \mathrm{Categorical}(w)$
        \Comment{past task $j$ chosen with prob.\ $w_j$}
      \State sample minibatch $b$ uniformly from
        $\mathcal{M}_j$
      \State \colorbox{yellow!25}{$\theta^{\mathrm{ref}} \gets
        \theta_j$}
        \Comment{\textsc{Reference}}
    \EndIf
    \State $\theta \gets \theta - \eta\, \nabla_\theta
      \big[\, \mathcal{L}_\mathrm{CE}(\theta; b) + \lambda\,
      \mathrm{KL}\big(f_{\theta^{\mathrm{ref}}}(b) \,\|\,
      f_\theta(b)\big) \,\big]$
  \EndFor
  \State $\mathcal{M}_t \gets$ subsample of $D_t$;\quad
    $\theta_t \gets \theta$;\quad
    $\mathcal{P} \gets \mathcal{P} \cup
    \{(g_t,\, \mathcal{M}_t,\, \theta_t)\}$
\EndFor
\end{algorithmic}
\end{algorithm}

\emph{Probe} --- the mean probe-batch gradient at the fixed
shared initialisation (Eq.~\ref{eq:signature};
Figure~\ref{fig:pipeline}a): the only tested signature that
reliably finds oracle-best sources, label-aware where
input-similarity signatures are not (\S\ref{sec:whom}); ten
forward--backward passes, no training.

\emph{Route} --- softmax over signature cosines at temperature
$\tau$ (Figure~\ref{fig:pipeline}b).  The temperature makes the
routing knob explicit: $\tau{\to}0$ recovers argmax selection,
$\tau{\to}\infty$ recovers uniform pooling ($=$ ER), and the
evaluated setting sits between the two extremes compared in
Appendix~\ref{app:variants}.

\emph{Replay} --- routed source data mixed into training at
ratio $\alpha$, batch by batch (Figure~\ref{fig:pipeline}c):
data is the persistent carrier of
\S\ref{sec:paramfails:carrier}, so the transferred signal
re-enters at every optimisation step rather than only at
initialisation.

\emph{Reference} --- each batch is distilled against the routed
record's era snapshot: $\theta_j$ on batches replayed from
$\mathcal{M}_j$, the top-routed $\theta_{j^\ast}$ on
current-task batches (Figure~\ref{fig:pipeline}c).  Why
distillation owns stability --- and why the teacher is an era
snapshot --- is the subject of \S\ref{sec:method:kd}.

Other instantiations of the three modules are configurations of
the same skeleton (Appendix~\ref{app:variants}).

\begin{figure}[t]
\centering
\begin{tikzpicture}[
  font=\small,
  box/.style={rectangle, draw, rounded corners, align=center,
              minimum height=0.9cm, inner sep=7pt},
  mem/.style={box, fill=blue!14, draw=blue!45},
  op/.style={box, fill=white},
  hot/.style={box, fill=orange!25, draw=orange!70!black, very thick},
  kd/.style={box, fill=green!14, draw=green!55!black},
  panel/.style={rounded corners=6pt, inner sep=10pt},
  plabel/.style={font=\bfseries, anchor=south west},
  arrow/.style={-{Stealth[round]}, very thick, gray!55!black},
  stream/.style={box, fill=gray!12, minimum width=1.2cm, minimum height=0.6cm},
  flow/.style={-{Stealth[round]}, line width=1.5pt, gray!55!black}
]
\node[stream] (t1) at (0,1.8) {$T_1$};
\node[stream, right=0.15cm of t1] (t2) {$T_2$};
\node[stream, right=0.15cm of t2] (td) {$\cdots$};
\node[stream, fill=orange!25, draw=orange!70!black, very thick,
      right=0.15cm of td] (tt) {$T_t$};
\node[right=0.3cm of tt, font=\small\itshape, text=gray]
  {Task stream $\rightarrow$};
\node[op, text width=5.4cm] (probe) at (0,0)
  {A few probe batches\\at the shared initialisation $\theta_0$};
\node[hot, text width=3.9cm, right=1.3cm of probe] (sig)
  {\textbf{Task signature} $g_t$};
\draw[arrow] (probe) -- (sig);
\node[mem, text width=5.4cm] (pool) at (0,-2.7)
  {\textbf{Task memory} $\mathcal{P}$: one record per past task\\
   signature $g_j$ \,\textbullet\, replay data $\mathcal{M}_j$
   \,\textbullet\, snapshot $\theta_j$};
\node[hot, text width=3.9cm, right=1.3cm of pool] (route)
  {\textbf{Match signatures}\\$\cos(g_j, g_t) \rightarrow w$\\picks data \emph{and} teacher};
\draw[arrow] (pool) -- (route);
\node[op, text width=5.4cm] (mix) at (0,-5.9)
  {\textbf{Replay} $b \sim \mathcal{M}_j$: routed source examples\\mixed into every training batch};
\node[kd, text width=3.9cm, right=1.3cm of mix] (kdn)
  {\textbf{Teacher} $\theta^{\mathrm{ref}} = \theta_j$:\\the routed record's snapshot};
\node[hot, minimum width=10.6cm,
      below=0.5cm of $(mix.south)!0.5!(kdn.south)$] (obj)
  {\textbf{Train}: $\mathcal{L}_{\mathrm{CE}} + \lambda\, \mathcal{L}_{\mathrm{KD}}$};
\draw[arrow] (mix.south) -- ([xshift=-2.3cm]obj.north);
\draw[arrow] (kdn.south) -- ([xshift=2.5cm]obj.north);
\begin{scope}[on background layer]
\node[panel, fill=orange!7, fit=(probe)(sig)] (pa) {};
\node[panel, fill=blue!6, fit=(pool)(route)] (pb) {};
\node[panel, fill=green!7, fit=(mix)(kdn)(obj)] (pc) {};
\end{scope}
\node[plabel] at ([yshift=2pt]pa.north west) {(a) Probe};
\node[plabel] at ([yshift=2pt]pb.north west) {(b) Route};
\node[plabel] at ([yshift=2pt]pc.north west) {(c) Train};
\draw[flow] (tt.south) -- (tt.south |- pa.north)
  node[midway, right, font=\small]{New task};
\draw[flow] (sig.south) -- (sig.south |- pb.north);
\draw[flow] (route.south) --
  node[midway, sloped, above=1pt, font=\small]{replay data $\mathcal{M}_j$}
  (mix.north east);
\draw[flow] (route.south) --
  node[midway, right=2pt, font=\small]{teacher $\theta_j$}
  (route.south |- kdn.north);
\end{tikzpicture}
\caption{\textbf{The \method{} pipeline}.}
\label{fig:pipeline}
\end{figure}

\subsection{Knowledge distillation for stability}
\label{sec:method:kd}

Once replay is reassigned to transfer, the stability term of
the continual objective is unowned; \methodshort{} assigns it
to distillation.  Each training batch $b$ is anchored by
\begin{equation}
\mathcal{L}_\mathrm{KD}(\theta;\, b) \;=\;
\mathrm{KL}\big(
f_{\theta^{\mathrm{ref}}}(b) \,\big\|\, f_\theta(b)
\big),
\label{eq:objective}
\end{equation}
added to the task loss with weight $\lambda$
(Algorithm~\ref{alg:tsrt}): the KL is taken between the
token-level predictive distributions of the frozen reference
and the student on the same batch, and the reference switches
with the batch's origin --- $\theta_j$ on batches replayed
from $\mathcal{M}_j$, the top-routed $\theta_{j^\ast}$ on
current-task batches.

The novelty is not the loss.  It
lies, first, in the \emph{role assignment}: replay serves
transfer, while distillation alone owns stability.  It lies,
second, in the \emph{binding}: teachers are queryable era
snapshots, routed jointly with the replay data through the task
memory, so one routing decision returns both what to replay and
whom to distil against, and the reference can answer on any
batch, including current-task ones.  To our
knowledge, distillation as an explicitly routed stability
module has no counterpart in recent continual learning with
LLMs.

\paragraph{What the teacher contributes.}  A systematic
teacher audit dissociates teacher \emph{knowledge} from
optimisation \emph{anchoring}: teacher identity is
second-order, the teacher's value lives in its base-shared
component, and every attempt to make distillation carry
transfer reduces both transfer \emph{and} stability.  In this
regime distillation functions as a stability regulariser ---
adjacent to reference-KL in RL~\citep{teh2017distral} and to
label-smoothing accounts of supervised
distillation~\citep{yuan2020revisiting,mobahi2020self}, though
the dissociation in continual LLM learning is, to our
knowledge, new --- and we assign it exclusively to stability.
The one teacher property that consistently matters in long
streams is \emph{drift}: the most recent snapshot is
increasingly biased toward the latest task, which is what
motivates the era-matched reference.  The full audit is
reported in Appendix~\ref{app:kd}.

\section{Experiments}\label{sec:experiments}

Sections~\ref{sec:transfer}--\ref{sec:method} measured \emph{why}
transfer should work and derived \methodshort{} as one
instantiation.  This section validates the framework under the
standard continual-learning protocol, in three questions: does
transfer improve low-budget continual learning
(\S\ref{sec:experiments:main}); what determines the size and
shape of the gains (\S\ref{sec:experiments:regime}); and how
much do the design choices matter
(\S\ref{sec:experiments:ablations}).

\subsection{Setup}\label{sec:experiments:setup}

\textbf{Models and streams.}  Qwen2.5-0.5B/7B-Instruct with a
single LoRA adapter~\citep{hu2021lora} trained continually ---
the dominant regime for continual learning with LLMs.  Two eight-task streams:
TRACE-8~\citep{wang2023trace} (heterogeneous: stance, finance,
summarisation, code, science QA, arithmetic) in its canonical
order, and NumGLUE-8~\citep{mishra2022numglue} (one
numerical-reasoning core in eight formats) in type order.

\textbf{Protocol.}  We follow the standard sequential protocol
of the systems we compare
against~\citep{wang2023olora,qian2025treelora,buzzega2020derpp}:
one accumulating adapter trains through all eight tasks;
after each task, all seen tasks are evaluated;
$n{=}5$ seeds per cell.  We evaluate in the low-budget
regime --- $N{=}50$ examples per task --- where transfer is
operationally interesting (\S\ref{sec:paramfails}); the
protocol structure is the literature's, the budget is the
studied setting.  All remaining configuration (adapter,
optimiser, training budgets, and \methodshort{}'s module
settings) is detailed in Appendix~\ref{app:setupdetail},
together with cost accounting and stream construction;
order-robustness checks are in Appendix~\ref{app:robust}.

\textbf{Metrics.}  Training uses token-level cross-entropy. Evaluation reports exact-match accuracy on classification tasks and evaluation loss on generation tasks, where exact match is degenerate. Let
$A_{k,i}$ denote accuracy on $T_i$'s test set after training on
task $T_k$.  We report the three standard sequential
metrics~\citep{lopezpaz2017gem, chaudhry2019agem}:
\begin{align}
\text{Overall} &= \frac{1}{K} \sum_{i=1}^{K} A_{K,i},
\label{eq:overall}\\
\text{Plas} &= \frac{1}{K} \sum_{i=1}^{K} A_{i,i},
\label{eq:plas}\\
\text{BWT} &= \frac{1}{K-1} \sum_{i=1}^{K-1}
  \left(A_{K,i} - A_{i,i}\right).
\label{eq:bwt}
\end{align}
Overall is the final average accuracy; plasticity (Plas) is
accuracy immediately after learning each task; and BWT is backward transfer, where $0$ indicates perfect retention, positive values indicate improvement on earlier tasks, and negative values indicate forgetting.

\textbf{Baselines.}  Naive sequential fine-tuning (SeqFT).
\emph{Rehearsal-free}: regularisation and distillation
(EWC~\citealp{kirkpatrick2017ewc}; LwF~\citealp{li2017lwf};
SDFT~\citealp{yang2024sdft}); parameter isolation
(O-LoRA~\citealp{wang2023olora}); and two NeurIPS'25 methods ---
GainLoRA~\citep{liang2025gainlora}, gated per-task LoRA branches
with projection-constrained gates, and DEAL~\citep{han2025deal},
wavelet-filtered retention on a continually edited adapter
(port details in Appendix~\ref{app:setupdetail}).
\emph{Rehearsal-based}: gradient constraints from memory samples
(A-GEM~\citealp{chaudhry2019agem});
DER++~\citealp{buzzega2020derpp}; and
ER~\citealp{chaudhry2019er} --- uniform replay over all past
tasks --- at a generous $50\%$ replay ratio, with a $10\%$
variant marking the budget spectrum.  Zero-shot exact match is $0$ under the
strict answer format and is therefore omitted from
Table~\ref{tab:method}.

\subsection{Does transfer improve low-budget continual
learning?}\label{sec:experiments:main}

We now test whether the transfer contribution of
\S\ref{sec:setup} improves continual learning under a fixed
target-data budget (Table~\ref{tab:method}).

\textbf{Plasticity.}  In the decomposition of
\S\ref{sec:setup}, plasticity is where the transfer
contribution should appear first, so we read Plas first.  On
TRACE, \methodshort{} posts the best plasticity at both scales
($.439$ at 0.5B, $.670$ at 7B): signature-weighted replay beats
ER on the transfer axis, the controlled probe's
finding carried into the stream.  On NumGLUE, methods that only
chase the current task can edge Plas (SDFT $.334$ at 0.5B,
SeqFT $.587$ at 7B) --- but they pay with the worst forgetting
in their columns; plasticity bought by abandoning the past is
not transfer.

\textbf{Overall accuracy.}  Overall
folds retention in, and \methodshort{} is best everywhere.  The
reading we care about is the budget one: with the same $50$
target examples per task, selected replay reaches final
performance that ER does not --- a gain that
would otherwise require additional target supervision, and that
Figure~\ref{fig:budget}(b) converts into equivalent target
labels on the probe (\S\ref{sec:paramfails}).  Stream-level
margins are smaller than probe-level ones for a predicted
reason: the stream itself gradually consumes pair-specific
headroom --- by the time a target arrives, its best source has
often already been trained into the adapter.

\textbf{Stability.}  The plasticity gains do not come at the
expense of forgetting: \methodshort{} is the only method in
Table~\ref{tab:method} with \emph{positive} BWT in all four
settings.  Replay incidentally protects the sources it draws
on, and distillation anchors every batch to an undrifted era
reference.  \textbf{Under a fixed target
budget, realised transfer buys higher plasticity and higher final
accuracy at no retention cost.}

\textbf{Backbone robustness.}  The same qualitative behaviour
is observed on two additional backbone families (Llama-3.2-1B
and Gemma-3-1B; Appendix~\ref{app:robust}): replay-based
methods remain strongest, and \methodshort{} again attains the
best overall accuracy, plasticity, and backward transfer ---
the framework's predictions are not specific to Qwen models.

\begin{table}[t]
\centering\tiny
\setlength{\tabcolsep}{1.1pt}
\resizebox{\textwidth}{!}{%
\begin{tabular}{l ccc ccc ccc ccc}
\toprule
 & \multicolumn{3}{c}{TRACE-8 (0.5B)} & \multicolumn{3}{c}{TRACE-8 (7B)} & \multicolumn{3}{c}{NumGLUE-8 (0.5B)} & \multicolumn{3}{c}{NumGLUE-8 (7B)} \\
\cmidrule(lr){2-4}\cmidrule(lr){5-7}\cmidrule(lr){8-10}\cmidrule(lr){11-13}
Method & Overall$\uparrow$ & Plas$\uparrow$ & BWT$\uparrow$ & Overall$\uparrow$ & Plas$\uparrow$ & BWT$\uparrow$ & Overall$\uparrow$ & Plas$\uparrow$ & BWT$\uparrow$ & Overall$\uparrow$ & Plas$\uparrow$ & BWT$\uparrow$ \\
\midrule
SeqFT
 & $.264{\scriptscriptstyle\pm .048}$ & $.398{\scriptscriptstyle\pm .015}$ & $-.134{\scriptscriptstyle\pm .058}$ & $.584{\scriptscriptstyle\pm .033}$ & $.660{\scriptscriptstyle\pm .001}$ & $-.076{\scriptscriptstyle\pm .031}$ & $.284{\scriptscriptstyle\pm .041}$ & $.316{\scriptscriptstyle\pm .018}$ & $-.037{\scriptscriptstyle\pm .053}$ & $.555{\scriptscriptstyle\pm .032}$ & $\mathbf{.587}{\scriptscriptstyle\pm .008}$ & $-.037{\scriptscriptstyle\pm .037}$ \\
\midrule
EWC
 & $.245{\scriptscriptstyle\pm .031}$ & $.427{\scriptscriptstyle\pm .004}$ & $-.182{\scriptscriptstyle\pm .031}$ & $.599{\scriptscriptstyle\pm .010}$ & $.666{\scriptscriptstyle\pm .011}$ & $-.067{\scriptscriptstyle\pm .014}$ & $.279{\scriptscriptstyle\pm .018}$ & $.320{\scriptscriptstyle\pm .014}$ & $-.047{\scriptscriptstyle\pm .030}$ & $.554{\scriptscriptstyle\pm .035}$ & $.584{\scriptscriptstyle\pm .009}$ & $-.034{\scriptscriptstyle\pm .034}$ \\
LwF
 & $.291{\scriptscriptstyle\pm .045}$ & $.424{\scriptscriptstyle\pm .013}$ & $-.133{\scriptscriptstyle\pm .033}$ & $.640{\scriptscriptstyle\pm .024}$ & $.668{\scriptscriptstyle\pm .001}$ & $-.029{\scriptscriptstyle\pm .023}$ & $.292{\scriptscriptstyle\pm .014}$ & $.313{\scriptscriptstyle\pm .013}$ & $-.024{\scriptscriptstyle\pm .008}$ & $.563{\scriptscriptstyle\pm .015}$ & $.580{\scriptscriptstyle\pm .017}$ & $-.020{\scriptscriptstyle\pm .007}$ \\
SDFT
 & $.305{\scriptscriptstyle\pm .070}$ & $.427{\scriptscriptstyle\pm .009}$ & $-.122{\scriptscriptstyle\pm .072}$ & $.605{\scriptscriptstyle\pm .011}$ & $.664{\scriptscriptstyle\pm .009}$ & $-.059{\scriptscriptstyle\pm .019}$ & $.279{\scriptscriptstyle\pm .033}$ & $\mathbf{.334}{\scriptscriptstyle\pm .016}$ & $-.063{\scriptscriptstyle\pm .027}$ & $.553{\scriptscriptstyle\pm .009}$ & $.580{\scriptscriptstyle\pm .011}$ & $-.030{\scriptscriptstyle\pm .011}$ \\
O-LoRA
 & $.233{\scriptscriptstyle\pm .028}$ & $.401{\scriptscriptstyle\pm .007}$ & $-.168{\scriptscriptstyle\pm .030}$ & $.617{\scriptscriptstyle\pm .010}$ & $.662{\scriptscriptstyle\pm .010}$ & $-.045{\scriptscriptstyle\pm .005}$ & $.281{\scriptscriptstyle\pm .016}$ & $.310{\scriptscriptstyle\pm .028}$ & $-.033{\scriptscriptstyle\pm .014}$ & $.567{\scriptscriptstyle\pm .006}$ & $.579{\scriptscriptstyle\pm .014}$ & $-.013{\scriptscriptstyle\pm .010}$ \\
GainLoRA
 & $.374{\scriptscriptstyle\pm .018}$ & $.398{\scriptscriptstyle\pm .022}$ & $-.024{\scriptscriptstyle\pm .005}$ & $.654{\scriptscriptstyle\pm .010}$ & $.663{\scriptscriptstyle\pm .009}$ & $-.009{\scriptscriptstyle\pm .005}$ & $.297{\scriptscriptstyle\pm .014}$ & $.307{\scriptscriptstyle\pm .017}$ & $-.011{\scriptscriptstyle\pm .004}$ & $.558{\scriptscriptstyle\pm .006}$ & $.563{\scriptscriptstyle\pm .005}$ & $-.005{\scriptscriptstyle\pm .004}$ \\
DEAL
 & $.372{\scriptscriptstyle\pm .019}$ & $.401{\scriptscriptstyle\pm .003}$ & $-.029{\scriptscriptstyle\pm .022}$ & $.224{\scriptscriptstyle\pm .153}$ & $.576{\scriptscriptstyle\pm .078}$ & $-.352{\scriptscriptstyle\pm .088}$ & $.253{\scriptscriptstyle\pm .024}$ & $.303{\scriptscriptstyle\pm .013}$ & $-.057{\scriptscriptstyle\pm .023}$ & $.418{\scriptscriptstyle\pm .023}$ & $.500{\scriptscriptstyle\pm .006}$ & $-.094{\scriptscriptstyle\pm .023}$ \\
\midrule
A-GEM
 & $.310{\scriptscriptstyle\pm .019}$ & $.418{\scriptscriptstyle\pm .016}$ & $-.107{\scriptscriptstyle\pm .021}$ & $.610{\scriptscriptstyle\pm .032}$ & $.662{\scriptscriptstyle\pm .007}$ & $-.052{\scriptscriptstyle\pm .025}$ & $.286{\scriptscriptstyle\pm .031}$ & $.325{\scriptscriptstyle\pm .005}$ & $-.045{\scriptscriptstyle\pm .037}$ & $.559{\scriptscriptstyle\pm .025}$ & $.586{\scriptscriptstyle\pm .014}$ & $-.031{\scriptscriptstyle\pm .013}$ \\
DER++
 & $.220{\scriptscriptstyle\pm .015}$ & $.262{\scriptscriptstyle\pm .022}$ & $-.042{\scriptscriptstyle\pm .037}$ & $.264{\scriptscriptstyle\pm .104}$ & $.288{\scriptscriptstyle\pm .096}$ & $-.023{\scriptscriptstyle\pm .081}$ & $.210{\scriptscriptstyle\pm .025}$ & $.212{\scriptscriptstyle\pm .028}$ & $-.003{\scriptscriptstyle\pm .007}$ & $.393{\scriptscriptstyle\pm .054}$ & $.445{\scriptscriptstyle\pm .010}$ & $-.060{\scriptscriptstyle\pm .051}$ \\
ER ($10\%$ replay)
 & $.406{\scriptscriptstyle\pm .019}$ & $.435{\scriptscriptstyle\pm .008}$ & $-.029{\scriptscriptstyle\pm .025}$ & $.638{\scriptscriptstyle\pm .029}$ & $.663{\scriptscriptstyle\pm .005}$ & $-.025{\scriptscriptstyle\pm .026}$ & $.321{\scriptscriptstyle\pm .020}$ & $.331{\scriptscriptstyle\pm .013}$ & $-.012{\scriptscriptstyle\pm .011}$ & $.563{\scriptscriptstyle\pm .011}$ & $.583{\scriptscriptstyle\pm .013}$ & $-.023{\scriptscriptstyle\pm .016}$ \\
ER
 & $.451{\scriptscriptstyle\pm .031}$ & $.427{\scriptscriptstyle\pm .024}$ & $+.025{\scriptscriptstyle\pm .020}$ & $.656{\scriptscriptstyle\pm .034}$ & $.662{\scriptscriptstyle\pm .018}$ & $-.006{\scriptscriptstyle\pm .046}$ & $.336{\scriptscriptstyle\pm .013}$ & $.314{\scriptscriptstyle\pm .011}$ & $+.025{\scriptscriptstyle\pm .010}$ & $.578{\scriptscriptstyle\pm .012}$ & $.579{\scriptscriptstyle\pm .015}$ & $-.001{\scriptscriptstyle\pm .019}$ \\
\midrule
\textbf{\methodshort{} (ours)}
 & $\mathbf{.475}{\scriptscriptstyle\pm .030}$ & $\mathbf{.439}{\scriptscriptstyle\pm .017}$ & $\mathbf{+.035}{\scriptscriptstyle\pm .015}$ & $\mathbf{.684}{\scriptscriptstyle\pm .008}$ & $\mathbf{.670}{\scriptscriptstyle\pm .012}$ & $\mathbf{+.014}{\scriptscriptstyle\pm .011}$ & $\mathbf{.353}{\scriptscriptstyle\pm .016}$ & $.330{\scriptscriptstyle\pm .012}$ & $\mathbf{+.026}{\scriptscriptstyle\pm .007}$ & $\mathbf{.588}{\scriptscriptstyle\pm .017}$ & $.582{\scriptscriptstyle\pm .012}$ & $\mathbf{+.007}{\scriptscriptstyle\pm .022}$ \\
\bottomrule
\end{tabular}}%

\caption{\textbf{Continual learning under the standard
sequential protocol} (TRACE-8 / NumGLUE-8 at 0.5B / 7B).  Mean
$\pm$ std over $5$ seeds; bold =
best per column.  Overall and Plas are \emph{absolute}
accuracies; BWT is the change from post-task to end-of-stream
accuracy; no row is normalised by the zero-shot reference
(omitted: exact match is $0$ under the strict answer format).
Rules separate the no-mechanism baseline, the rehearsal-free
family, the rehearsal-based family, and ours.
Overall, Plas and BWT as in
Eqs.~\eqref{eq:overall}--\eqref{eq:bwt}; baselines and protocol
in \S\ref{sec:experiments:setup}; routing-policy variants in
Appendix~\ref{app:variants}; per-task results in
Appendix~\ref{app:pertask}.}
\label{tab:method}
\end{table}

\subsection{What determines transfer?}
\label{sec:experiments:regime}
\label{sec:experiments:kd}
\label{sec:experiments:why}
\label{sec:experiments:numglue}
\label{sec:experiments:7b}

The framework predicts that realised transfer depends jointly
on transfer opportunity and task structure --- and indeed the
same method, protocol, and budget produce gains of very
different size and shape across the four settings of
Table~\ref{tab:method}.  Our evaluation scope is deliberately
\textbf{two-dimensional --- target-data budget and source-task
structure} --- the same
axes along which the reuse of previously collected data is
evaluated in online RL~\citep{ball2023efficient}.  The
differences follow the framework, not the method: two
observations and two boundaries.

\paragraph{Transfer follows task structure.}  TRACE is
heterogeneous: sources differ sharply and several hurt
(Table~\ref{tab:matrix}), so \emph{selection} is what pays ---
routing beats pooling on plasticity, and argmax selection is
strongest on the pure-transfer probe
(Appendix~\ref{app:jointtablesec}).  NumGLUE is the opposite
regime --- one arithmetic core in eight formats: headroom is
large and dense (Appendix~\ref{app:numglue}), every source
carries the core, and within such homogeneous families the
framework predicts \emph{pooling} rather than single-source
replay, which the controlled probe confirms.  The signature's
resolution also drops inside a family (its cosines compress;
Appendix~\ref{app:ngmatrix}) --- a bound on the current
instantiation, not on the framework, since the selection module
is the replaceable one (\S\ref{sec:method:alg}).  The routed
triple stays best in the stream in both regimes because its
soft routing spans them: sharp enough to select on TRACE, broad
enough to keep coverage on NumGLUE.

\paragraph{Transfer follows headroom.}
Figure~\ref{fig:scale} repeats the headroom audit across
Qwen2.5 scales on TRACE: as the base grows, the target-data
arm saturates on its own and the best-source gain decays
monotonically to zero; at 7B the oracle buys nothing.  This is
a probe-level statement
about opportunity, not a method comparison, and it agrees with
Table~\ref{tab:method}: every TRACE Plas column flattens at 7B
--- no policy can transfer where no opportunity remains ---
while Overall still separates through retention.

\begin{figure}[t]
\centering
\begin{tikzpicture}
\begin{axis}[
  width=0.58\linewidth, height=4.3cm,
  symbolic x coords={0.5B,1.5B,3B,7B}, xtick=data,
  xlabel={Base-model scale},
  axis y line*=left,
  ymin=0.40, ymax=0.72,
  ylabel={Target-data accuracy},
  ylabel style={blue!55!black}, y tick label style={blue!55!black},
  tick label style={font=\scriptsize},
  label style={font=\scriptsize},
]
\addplot[blue!55!black, thick, mark=*, mark size=1.4pt]
  coordinates {(0.5B,0.447) (1.5B,0.566) (3B,0.611) (7B,0.668)};
\draw[-{Stealth[round]}, thick, dashed, blue!55!black]
  (axis cs:3B,0.648) -- ++(0.75cm,0.18cm);
\end{axis}
\begin{axis}[
  width=0.58\linewidth, height=4.3cm,
  symbolic x coords={0.5B,1.5B,3B,7B}, xtick=data,
  axis y line*=right, axis x line=none,
  ymin=-0.004, ymax=0.019,
  ylabel={Best-source gain},
  ylabel style={green!40!black}, y tick label style={green!40!black},
  tick label style={font=\scriptsize},
  label style={font=\scriptsize},
  scaled y ticks=false,
  yticklabel style={/pgf/number format/fixed},
]
\addplot[green!55!black, very thick, mark=square*, mark size=1.4pt]
  coordinates {(0.5B,0.015) (1.5B,0.011) (3B,0.003) (7B,0.000)};
\draw[-{Stealth[round]}, thick, dashed, green!55!black]
  (axis cs:3B,0.006) -- ++(0.8cm,-0.28cm);
\end{axis}
\end{tikzpicture}
\caption{\textbf{Transfer follows headroom across model scale}
(TRACE headroom audit per Qwen2.5 scale; error bands omitted
for clarity).}
\label{fig:scale}
\end{figure}
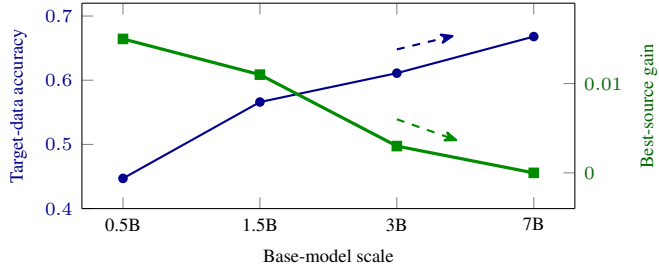  NumGLUE shows the graded version of the same law: its
in-domain headroom persists at 7B, but the fraction other tasks
can supply compresses ($+0.033 \to +0.003$) --- scale erodes
what tasks share before it erodes what each still individually
lacks.  Notably, signature picks are scale-stable (7B agrees
with 0.5B on $7/8$ targets), so signatures can be computed once
on a small proxy model.

\paragraph{Boundaries.}  The framework predicts its own two
edges, and both are observed: \emph{saturation} --- on
benchmarks without headroom, transfer is unmeasurable by any
mechanism; and \emph{generic transfer} --- the regime where most sources
provide similar benefit, so source identity matters little: on
six Super-NaturalInstructions pairs headroom exists and source
data helps, but sources are near-interchangeable and the value
of selection is small (Appendix~\ref{app:superni}).
\textbf{Thus, task structure determines how transfer should be
realised, whereas headroom determines whether it can be
realised at all.}

\subsection{Design choices}
\label{sec:experiments:ablations}

The framework specifies \emph{what} must be decided, not
\emph{how}: Algorithm~\ref{alg:tsrt} leaves implementation
choices open, and we ablate these --- how replay is routed, and
whether task order still matters once routing is available.

\paragraph{Routing.}  The routing module
decides whether replay comes from a single selected source or a
weighted combination; the framework is agnostic, so we compare
argmax selection, soft routing, and uniform pooling.  In the
stream, soft routing --- the routed triple --- beats both
extremes and its single-component variants
(Table~\ref{tab:method}; Appendix~\ref{app:variants}); on the
controlled probe, even a softmax-weighted top-3 mixture
\emph{lowers} mean accuracy against the single selected
source.  The reason is
the shape of the matrix: on heterogeneous streams it is sparse
--- the second- and third-best sources are often weak or
harmful, and averaging them dilutes transfer.  \textbf{Sparse
transfer matrices favour selection over averaging.}

\paragraph{Task order.}  If the matrix is
usable prospectively, ordering tasks so strong sources precede
their targets should help.  We compare the canonical order, the
greedy best-source curriculum (fixed before evaluation), its
reversal, and arbitrary reshuffles, at both scales
(Appendix~\ref{app:robust}).  The replay family moves by at
most $.03$ under any of them, on both streams and both scales:
replay already delivers the relevant source at training time,
so a curriculum has little left to contribute.
\textbf{Selection replaces curriculum.}

\section{Discussion}\label{sec:discussion}

\subsection{Implications}

This work does not propose another replay algorithm; it changes
the questions asked before one is designed.  Three reversals
follow.  \emph{Replay}: long treated as a forgetting-mitigation
tool, replay is the cheapest persistent carrier of transfer ---
pointing it at the new task turns a stability mechanism into a
plasticity mechanism, and distillation absorbs the stability
role it leaves behind.  \emph{Transfer}: what looked like a
parameter-reuse problem is a data-selection problem --- the
transferred signal must survive optimisation, and among
persistent carriers data is the cheapest --- so the productive
question is not how to merge parameters but which examples to
select.  \emph{Benchmarks}: a continual-learning benchmark
measures more than accuracy --- its headroom and transfer
structure decide whether transfer is measurable on it at all,
and both belong in its specification, alongside whether the
task order is canonical, arbitrary, or optimised; selection
makes replay-based learners order-insensitive while order
dominates learners without data access
(\S\ref{sec:experiments:ablations}), so curricula and selection
are substitute uses of the same measured matrix.

\subsection{Limitations and open questions}

The framework was validated in the low-budget regime, where
transfer opportunity is measurable; whether it yields equally
useful mechanisms under richer supervision is an empirical
question.  Within this regime, each limitation pairs with the
question it opens.

\emph{Storage.}  Replay assumes access to past task data.  The
buffer is small (${\sim}100$ examples per task, ${\sim}3$\,MB
for $15$ tasks, favourable versus per-task adapter libraries),
but privacy-constrained deployments cannot store raw examples
--- opening the question of whether transfer can be
\emph{carried differently}: synthetic or distilled replay and
retrieval are candidates, judged by the same persistence
criterion (\S\ref{sec:paramfails:carrier}).

\emph{The current instantiation.}  The implementation
instantiates the framework with one task signature and one
replay policy; neither is claimed optimal.  Signature
resolution drops within homogeneous families
(\S\ref{sec:experiments:numglue}) --- opening the question of
whether transfer can be \emph{predicted better}: task
signatures that retain within-family resolution, and signatures
for modalities beyond text.

\emph{Evaluation.}  Existing benchmarks are designed to expose
forgetting, not transfer --- headroom had to be audited here
before any mechanism could be tested --- opening the question
of whether transfer can be \emph{benchmarked directly}: streams
with reported headroom, a published pairwise matrix, and known
negative cells, so that transfer is evaluated as directly as
forgetting is today.

Beyond language, the conditions are stated at the level of
tasks and data: vision and multimodal streams, and
budget-allocation problems outside supervised language
modelling --- e.g., allocating simulation budget for scientific
surrogates~\citep{meng2026rlmesh} --- pose the same
fixed-budget transfer question.

\subsection{Conclusion}

\emph{Transfer in continual learning is a data-selection
problem rather than a parameter-reuse problem.}  The two
carrier studies point at one law: continual optimisation
preserves persistent signals and suppresses transient,
task-specific ones.
\methodshort{} is one instantiation; the framework is intended
to outlive it.  The particular signature, replay policy, and
stability mechanism are all replaceable --- what we hope
remains is the discipline: before proposing a transfer
mechanism, first ask whether transfer is possible, what can
carry it, and whom it should come from.  We hope continual
learning comes to evaluate transfer as systematically as it has
evaluated forgetting.

\section{acknowledgments}
We gratefully acknowledge the support of NSF IIS-2313131, NSF IIS-2332475 and NSF IIS-2543755, the NSF-Simons AI-Institute for the Sky (SkAI) via grants
NSF AST-2421845 and Simons Foundation MPS-AI00010513, and the University of Chicago’s Research Computing Center for their support of this work.
\bibliographystyle{iclr2026_conference}
\bibliography{main}

\appendix
\clearpage
\section*{Appendix}

\noindent\textbf{Appendix guide.}
Appendix~\ref{app:benchmarks} collects the benchmark-level
evidence for the framework of \S\ref{sec:transfer}: the four
regimes --- PP15 (no headroom), TRACE (controlled transfer
measurement), SuperNI (generic transfer), and NumGLUE
(homogeneous transfer).
Appendix~\ref{app:implementation} collects the implementation
evidence behind \S\ref{sec:method}: carrier, replay, routing,
and distillation.
Appendix~\ref{app:details} is the experimental record behind
\S\ref{sec:experiments}: setup, per-task results, robustness,
and reproducibility.  Unless stated otherwise, appendix results
follow the probe protocol of \S\ref{sec:paramfails} and the
stream protocol of \S\ref{sec:experiments:setup}.

\section{Benchmark Regimes}\label{app:benchmarks}

\subsection{PP15: the zero-headroom regime}\label{app:pp15}

Under the audit of \S\ref{sec:paramfails}, PP15 with an
instruction-tuned base leaves essentially no transfer
opportunity: ten times more target supervision improves
accuracy by only $+0.02$ on average.  Consistently, the best of
fourteen parameter-reuse configurations (oracle-selected
source; initialisation, perturbation, moment, and projection
families; $n{=}5$) improves the no-transfer baseline by
$+0.005$ --- within one seed standard deviation --- and the
mean across configurations is $-0.008$.  No transient advantage
survives past ${\sim}50$ steps under early-budget checks.

\paragraph{Protocol caution.}  An earlier oracle sweep under an
asymmetric merge protocol reported peer-pair gains up to
$+0.96$; controlled re-measurement attributed ${\sim}97\%$ of
that signal to an eval-path artefact of the merge rule ---
oracle definitions must be protocol-clean.

\subsection{TRACE-8: controlled transfer measurement}\label{app:jointtablesec}

This appendix records the controlled protocol behind the
transfer claims of \S\ref{sec:transfer}: the five probe arms
(target-data, target-full, mixed with a chosen source, random
and hindsight-oracle selection) run per target on a fresh
adapter.

\paragraph{Budget sweep.}  The sweep behind
Figure~\ref{fig:budget} runs every arm at
$N \in \{10, 20, 50, 100, 200, 500\}$ on the five TRACE
classification targets (sources always contribute $500$
examples; mixing $50/50$).  Three regularities hold across
targets and confirm the flagship reading: the opportunity
shrinks monotonically as $N$ grows; the selected-replay gain
tracks it and concentrates on matrix-strong targets ---
numglue-cm captures $35\%$ of its remaining opportunity at
$N{=}10$ ($+.066$, a $2\times$ label efficiency) --- while
matrix-weak targets sit near zero and pay for the forced
mixture once saturated; and the parameter-init arm never
separates from zero at any budget.

\paragraph{Task-signature robustness.}\label{app:signature}
Each signature uses $10$ probe batches and is scored by
best-source top-1 hit rate against the measured matrix (chance
${\approx}14\%$, i.e.\ $1/7$).  Two observations complement
Table~\ref{tab:predictors}.
\emph{The signal is distributed across depth}: restricting the
gradient signature to the last four blocks
(\texttt{GradB\_last4}) drops top-1 from $4/8$ to $3/8$, so no
single depth carries the alignment.  \emph{The fixed shared
initialisation matters}: every task's gradient is taken at the
same reference point, so the signatures form a stable,
training-free pairwise object --- unlike gradient similarities
computed at evolving weights during CL (e.g., for adapter
routing~\citep{qian2025treelora}).

\paragraph{Local intervention.}\label{app:intervention}
\emph{Protocol.}  Three classification targets spanning the
structure of Table~\ref{tab:matrix} (C-STANCE: weak, mixed
structure; ScienceQA: negative sources present; NumGLUE-ds:
strong positive transfer).  Each repetition draws three
independent batches --- one for the target gradient, one for a
source gradient, one held-out target batch for evaluation ---
and applies pure parameter perturbations (no optimiser state),
all normalised to the same norm $\epsilon$: the
\emph{source-only} effect is
$L_t(\theta-\epsilon\hat g_s)-L_t(\theta)$, and the
\emph{marginal} effect adds the source step after one
equal-norm target step $\theta_T$, i.e.\
$L_t(\theta_T-\epsilon\hat g_s)-L_t(\theta_T)$; a second,
independent target batch added to $\theta_T$ gives the
equal-compute reference.  To first order the source-only
effect is $-\epsilon\,\nabla L_t(\theta)^{\!\top}\hat g_s$, so
alignment should control immediate descent.  We use
$\epsilon\in\{0.25,0.5,1\}\times$ the measured one-step
parameter displacement and $30$ repetitions per state.
\emph{Results.}  At the shared initialisation, the
within-target rank correlation between gradient cosine and
immediate loss reduction is $\rho=.93$ (source-only) and
$.87$ (marginal) over $90$ repetitions.  The group ordering of
Table~\ref{tab:predictors} is strict at all three scales, with
helpful-step rates of $100\%/96\%/67\%$ for
top/middle/bottom sources and the bottom pair turning harmful
at the largest scale.  The stored signature preserves the
current gradient geometry (Spearman $.85$), so the quantity
\methodshort{} routes on identifies the sources whose
gradients provide compatible descent; the top-aligned pair
recovers roughly $28\%$ of the descent an additional target
batch would provide, without additional target labels.
\emph{Scope.}  The intervention is diagnostic at the shared
initialisation, where \methodshort{} measures its signatures.
Later checkpoints were also examined; the one-step probe loses
power once the $50$-example target budget is largely fitted,
including for target gradients themselves.  We therefore scope
the claim to the shared initialisation; long-horizon effects
are what the stream experiments measure.

\subsection{SuperNI: the generic-transfer regime}\label{app:superni}

SuperNI illustrates the generic-transfer boundary predicted by
the framework (\S\ref{sec:experiments:regime}).  On six
held-out Super-NaturalInstructions pairs ($n{=}5$), headroom
exists: additional in-domain supervision improves loss on
$6/6$ pairs --- PP15's zero-headroom regime, not TRACE's, is
the anomaly.  Source data helps: mixed training improves target
loss on $5/6$ pairs.  But selection has almost no value:
related and unrelated sources help nearly equally ($+0.04$ loss
difference).  The selection premium collapses exactly where
source-quality variance does; we report this boundary rather
than average over it.

\subsection{NumGLUE-8: the homogeneous regime}

\subsubsection{Headroom audit}\label{app:numglue}

NumGLUE-8 casts the eight task types of the NumGLUE
suite~\citep{mishra2022numglue} as the homogeneous regime: one
numerical-reasoning core in eight formats (types~4/5 are
TRACE's numglue-cm/ds).  The \S\ref{sec:setup:measure} audit
gives three regularities.  Headroom exists and persists across
scale: $7/8$ types retain accuracy headroom at 0.5B and at 7B
($n{=}5$).  Pooling beats single-source selection on the probe
at both scales ($0.348$ vs.\ $0.342$ at 0.5B, $0.600$ vs.\
$0.585$ at 7B) --- the pooling prediction of \S\ref{sec:whom}.
And the within-family oracle ceiling compresses from $+0.033$
to $+0.003$ across the same scales: in-domain headroom persists
while the share other tasks can supply shrinks.

\subsubsection{Pairwise transfer matrix}\label{app:ngmatrix}

Full transfer matrix ($\Delta$acc) behind
\S\ref{sec:experiments:numglue} ($\Delta$acc of $50/50$ joint
training vs.\ target-data; $n{=}5$; green = helps, red = hurts,
bold = column best; T4/T5 are TRACE's numglue-cm/ds).

\begin{table}[h]
\centering\small
\begin{tabular}{l|cccccccc}
\toprule
src$\backslash$tgt & T1 & T2 & T3 & T4 & T5 & T6 & T7 & T8 \\
\midrule
T1 & --- & \cellcolor{red!12}$-0.013$ & \cellcolor{red!25}$-0.015$ & \cellcolor{green!25}$+0.015$ & \cellcolor{green!10}$+0.008$ & \cellcolor{red!25}$-0.015$ & \cellcolor{red!25}$-0.018$ & \cellcolor{green!10}$+0.006$ \\
T2 & \cellcolor{green!45}$\mathbf{+0.060}$ & --- & \cellcolor{red!25}$-0.015$ & \cellcolor{green!45}$+0.041$ & \cellcolor{green!25}$+0.016$ & \cellcolor{red!12}$-0.011$ & \cellcolor{gray!6}$+0.004$ & \cellcolor{green!25}$+0.017$ \\
T3 & \cellcolor{green!10}$+0.012$ & \cellcolor{green!10}$+0.006$ & --- & \cellcolor{green!10}$+0.013$ & \cellcolor{gray!6}$-0.003$ & \cellcolor{gray!6}$-0.002$ & \cellcolor{green!10}$\mathbf{+0.005}$ & \cellcolor{red!12}$-0.006$ \\
T4 & \cellcolor{green!45}$+0.054$ & \cellcolor{gray!6}$+0.001$ & \cellcolor{gray!6}$-0.003$ & --- & \cellcolor{green!45}$\mathbf{+0.041}$ & \cellcolor{red!12}$-0.013$ & \cellcolor{red!25}$-0.015$ & \cellcolor{green!45}$\mathbf{+0.047}$ \\
T5 & \cellcolor{green!10}$+0.012$ & \cellcolor{red!12}$-0.007$ & \cellcolor{red!25}$-0.027$ & \cellcolor{green!25}$+0.037$ & --- & \cellcolor{red!25}$-0.024$ & \cellcolor{red!25}$-0.026$ & \cellcolor{red!12}$-0.009$ \\
T6 & \cellcolor{green!25}$+0.024$ & \cellcolor{green!25}$\mathbf{+0.015}$ & \cellcolor{red!12}$-0.006$ & \cellcolor{gray!6}$+0.000$ & \cellcolor{green!10}$+0.007$ & --- & \cellcolor{red!12}$-0.005$ & \cellcolor{green!10}$+0.009$ \\
T7 & \cellcolor{green!45}$+0.048$ & \cellcolor{red!12}$-0.012$ & \cellcolor{green!25}$\mathbf{+0.018}$ & \cellcolor{green!45}$+0.041$ & \cellcolor{green!10}$+0.011$ & \cellcolor{red!12}$-0.007$ & --- & \cellcolor{red!12}$-0.009$ \\
T8 & \cellcolor{green!45}$+0.054$ & \cellcolor{red!12}$-0.009$ & \cellcolor{red!25}$-0.030$ & \cellcolor{green!45}$\mathbf{+0.048}$ & \cellcolor{green!10}$+0.009$ & \cellcolor{green!10}$\mathbf{+0.005}$ & \cellcolor{red!25}$-0.017$ & --- \\
\bottomrule
\end{tabular}
\caption{Pairwise transfer matrix on NumGLUE-8.}
\label{tab:ngmatrixfull}
\end{table}

\section{Implementation Evidence}\label{app:implementation}

\subsection{The carrier}\label{app:scope}

Our evidence rules out
parameter reuse \emph{as transient optimiser state}:
initialising from, perturbing toward, or injecting the moments
of a past adapter into a module that then continues training ---
the form parameter reuse takes in dominant LLM-CL practice.  It
does \emph{not} contradict the two families of parameter-side
successes in the literature, and in fact explains both.
Inference-time composition is never trained against --- nothing
erases it.  And architectural sharing in the
Progressive-Networks / CAT / CTR lineage keeps past parameters
\emph{frozen and present at every forward pass} --- a persistent
signal, like data, rather than a one-shot initial condition.
The unifying property is persistence: \emph{signals that
participate in every training step survive; one-shot parameter
states are washed out}.  The property grades, rather than
merely splits, the mechanism space: an L2 anchor toward the
source adapter --- a \emph{persistent parameter} signal ---
captures roughly two-thirds of data mixing's gain at the
better anchor strength where one-shot initialisation captures
none (mean $0.4526$ vs.\ $0.4472$
target-data and $0.4547$ data; Appendix~\ref{app:ablations}).  The three
carriers therefore occupy the persistence axis exactly as
predicted.

\paragraph{A structural intuition.}  LoRA's update $BA$ is a
\emph{residual} on the shared base: it encodes what remains of a
task after the base has accounted for what the tasks share.  Two
related tasks can have near-orthogonal residuals (the shared
part was subtracted), which is why parameter similarity
under-determines transfer.  What related tasks actually share
--- the capability the base lacks --- lives in their data, and
is re-suppliable at every step.

\subsection{Replay and anchor ablations}\label{app:ablations}

\paragraph{Persistent parameter anchoring (L2-toward-source).}
Training the target with $L = \mathrm{CE} + \mu \lVert \theta -
\theta_\mathrm{source} \rVert^2$ keeps the source adapter present
at every step --- a persistent signal whose carrier is
\emph{parameters}.  On the five oracle pairs ($n{=}5$):
mean accuracy $0.4526$ at $\mu{=}0.1$ and $0.4410$ at
$\mu{=}1.0$, vs.\ $0.4472$ target-data and $0.4547$ for data
mixing.  Persistence is necessary (the one-shot init captures
nothing) but not sufficient: the data carrier remains stronger
at every $\mu$ we probed.

\subsection{Routing-policy variants}

\label{app:variants}

The routed triple has two knobs: the routing temperature $\tau$
(argmax at $\tau{\to}0$, uniform pooling at
$\tau{\to}\infty$) and the teacher binding (recent snapshot
vs.\ the routed batch's era snapshot).  All single-component
variants, same protocol and budgets as Table~\ref{tab:method}:

\begin{table}[h]
\centering\small
\resizebox{\textwidth}{!}{%
\begin{tabular}{l ccc ccc ccc ccc}
\toprule
 & \multicolumn{3}{c}{TRACE-8 (0.5B)} & \multicolumn{3}{c}{TRACE-8 (7B)} & \multicolumn{3}{c}{NumGLUE-8 (0.5B)} & \multicolumn{3}{c}{NumGLUE-8 (7B)} \\
\cmidrule(lr){2-4}\cmidrule(lr){5-7}\cmidrule(lr){8-10}\cmidrule(lr){11-13}
Method & Overall & Plas & BWT & Overall & Plas & BWT & Overall & Plas & BWT & Overall & Plas & BWT \\
\midrule
ER (uniform pooling; $\tau{\to}\infty$)
 & $.451{\scriptscriptstyle\pm .031}$ & $.427{\scriptscriptstyle\pm .024}$ & $+.025{\scriptscriptstyle\pm .020}$ & $.656{\scriptscriptstyle\pm .034}$ & $.662{\scriptscriptstyle\pm .018}$ & $-.006{\scriptscriptstyle\pm .046}$ & $.336{\scriptscriptstyle\pm .013}$ & $.314{\scriptscriptstyle\pm .011}$ & $+.025{\scriptscriptstyle\pm .010}$ & $.578{\scriptscriptstyle\pm .012}$ & $.579{\scriptscriptstyle\pm .015}$ & $-.001{\scriptscriptstyle\pm .019}$ \\
argmax selection ($\tau{\to}0$, recent teacher)
 & $.439{\scriptscriptstyle\pm .033}$ & $.430{\scriptscriptstyle\pm .014}$ & $+.010{\scriptscriptstyle\pm .019}$ & $.675{\scriptscriptstyle\pm .010}$ & $.668{\scriptscriptstyle\pm .015}$ & $+.007{\scriptscriptstyle\pm .009}$ & $.308{\scriptscriptstyle\pm .013}$ & $.308{\scriptscriptstyle\pm .012}$ & $+.000{\scriptscriptstyle\pm .016}$ & $.572{\scriptscriptstyle\pm .011}$ & $.577{\scriptscriptstyle\pm .011}$ & $-.005{\scriptscriptstyle\pm .007}$ \\
soft routing (recent teacher)
 & $.459{\scriptscriptstyle\pm .011}$ & $.436{\scriptscriptstyle\pm .015}$ & $+.023{\scriptscriptstyle\pm .006}$ & $.664{\scriptscriptstyle\pm .026}$ & $.668{\scriptscriptstyle\pm .010}$ & $-.004{\scriptscriptstyle\pm .028}$ & $.338{\scriptscriptstyle\pm .010}$ & $.316{\scriptscriptstyle\pm .011}$ & $+.025{\scriptscriptstyle\pm .018}$ & $.572{\scriptscriptstyle\pm .014}$ & $.577{\scriptscriptstyle\pm .012}$ & $-.005{\scriptscriptstyle\pm .026}$ \\
pool-hybrid (recent teacher)
 & $.459{\scriptscriptstyle\pm .020}$ & $.435{\scriptscriptstyle\pm .007}$ & $+.024{\scriptscriptstyle\pm .014}$ & $.669{\scriptscriptstyle\pm .014}$ & $.670{\scriptscriptstyle\pm .017}$ & $-.001{\scriptscriptstyle\pm .019}$ & $.333{\scriptscriptstyle\pm .009}$ & $.318{\scriptscriptstyle\pm .014}$ & $+.018{\scriptscriptstyle\pm .015}$ & $.566{\scriptscriptstyle\pm .040}$ & $.575{\scriptscriptstyle\pm .016}$ & $-.010{\scriptscriptstyle\pm .030}$ \\
\textbf{routed triple (full \methodshort{})}
 & $\mathbf{.475}{\scriptscriptstyle\pm .030}$ & $\mathbf{.439}{\scriptscriptstyle\pm .017}$ & $\mathbf{+.035}{\scriptscriptstyle\pm .015}$ & $\mathbf{.684}{\scriptscriptstyle\pm .008}$ & $\mathbf{.670}{\scriptscriptstyle\pm .012}$ & $\mathbf{+.014}{\scriptscriptstyle\pm .011}$ & $\mathbf{.353}{\scriptscriptstyle\pm .016}$ & $\mathbf{.330}{\scriptscriptstyle\pm .012}$ & $\mathbf{+.026}{\scriptscriptstyle\pm .007}$ & $\mathbf{.588}{\scriptscriptstyle\pm .017}$ & $\mathbf{.582}{\scriptscriptstyle\pm .012}$ & $\mathbf{+.007}{\scriptscriptstyle\pm .022}$ \\
\bottomrule
\end{tabular}}%
\caption{Routing-policy variants of \methodshort{} under the standard stream protocol.}
\label{tab:variantsfull}
\end{table}

Each ablated variant loses on the axis its missing
component serves (argmax loses coverage, pooling loses
selection, recent teachers lose the undrifted reference).

\subsection{Distillation: the full ablation record}\label{app:kd}

This appendix records the distillation audit referenced in
\S\ref{sec:experiments:kd}.  All experiments: PP15, sequential CL with
a single shared LoRA, $q,k,v,o$ projections, $500$ steps/task,
$n{=}5$ seeds, final mean accuracy (AVG) unless noted.  The
cumulative conclusion: KD in this setting is a trajectory
regulariser anchored to the shared base --- a stability mechanism
--- and cannot be converted into a transfer channel by teacher
choice, loss reshaping, timing, weighting, or snapshot source.

\subsubsection{Teacher identity is irrelevant (routing
  ceiling)}\label{app:kd:routing}

If KD transferred teacher-specific knowledge, teacher choice
should matter.  It does not:

\begin{table}[h]
\centering\small
\begin{tabular}{lcc}
\toprule
Teacher selection & AVG & note \\
\midrule
cosine prototype (default)       & $0.724 \pm 0.018$ & \\
random teacher per batch         & $0.715 \pm 0.022$ & ties cosine \\
per-batch oracle (min-CE teacher)& $0.715 \pm 0.004$ & \emph{ceiling}; ties random \\
latest snapshot only             & $0.703 \pm 0.001$ & systematically biased, $-0.02$ \\
anchor to frozen base            & $0.659 \pm 0.008$ & no snapshot at all \\
\midrule
no distillation (sequential FT)  & $0.584 \pm 0.022$ & \\
\bottomrule
\end{tabular}
\caption{Teacher-identity ablation for the distillation audit.}
\label{tab:kdteacherid}
\end{table}

Even the per-batch oracle --- the tightest upper bound on routing
benefit --- adds nothing over a random teacher.  The gain over no
distillation ($+0.13$) is therefore attributable to having
\emph{an} anchor, not to choosing the right one.

\subsubsection{Six loss-reshaping interventions all
  fail}\label{app:kd:falsification}

Each intervention was designed to amplify or isolate
teacher-specific signal; every one hurts:

\begin{table}[h]
\centering\small
\begin{tabular}{lcc}
\toprule
Variant & AVG & vs.\ plain KD ($0.724$) \\
\midrule
amplify teacher LoRA scale ($\gamma{=}4$)          & $0.493$ & $-0.231$ \\
distill on residual (teacher $-$ base)             & $0.516$ & $-0.208$ \\
push student away from base ($\lambda{=}10^{-3}$)  & $0.234$ & $-0.490$ \\
selective KL (gate by teacher correctness)         & $0.602$ & $-0.122$ \\
mixup CE                                           & collapse & --- \\
struggle-gate / margin transfer / hidden-MSE       & $0.638 / 0.600 / 0.681$ & all negative \\
\bottomrule
\end{tabular}
\caption{Loss-reshaping interventions on the distillation term.}
\label{tab:kdreshape}
\end{table}

Removing the base-shared component or repelling the student from
the base is catastrophic --- direct evidence that the value of KD
lies in the base-shared anchor, not in task-specific teacher
content.

\subsubsection{Timing and weight: regulariser
  signatures}\label{app:kd:timing}

KD interleaved with CE beats KD applied as a post-hoc phase
(endpoint correction), across two initialisation modes:

\begin{table}[h]
\centering\small
\begin{tabular}{lcc}
\toprule
Init & KD during CE & KD after CE \\
\midrule
QR          & $0.673 \pm 0.015$ & $0.651 \pm 0.010$ \\
routed mix  & $0.627 \pm 0.016$ & $0.614 \pm 0.023$ \\
\bottomrule
\end{tabular}
\caption{Distillation timing: interleaved versus post-hoc.}
\label{tab:kdtiming}
\end{table}

The weight curve is shallow over an order of magnitude (AVG
$0.673$--$0.683$ for $\lambda \in [0.1, 2]$) --- no narrow
optimum, no collapse at either end: the canonical signature of
a soft regulariser.

\subsubsection{Snapshot source is irrelevant}\label{app:kd:source}

Anchoring to an accumulated merged snapshot versus a routed
per-task snapshot makes no measurable difference (QR init:
$0.673$ vs.\ $0.673$; routed-mix init: $0.627$ vs.\ $0.607$,
within $1\sigma$): both pull the student toward the same
base-shared manifold.

\subsubsection{Transfer channels through distillation:
  local signal exists, none survives}\label{app:kd:channels}

Beyond loss reshaping, we tested whether snapshots carry
extractable transfer along three channels, plus a
vision-imported learnable bridge.  A zero-step diagnostic first
confirmed a real \emph{local} signal: perturbing parameters along
a snapshot's update direction predicts loss change with Pearson
$r = 0.93$ at small step sizes.  None of it survives training:

\begin{table}[h]
\centering\small
\begin{tabular}{llc}
\toprule
Channel & Method & best AVG \\
\midrule
logit     & gated multi-teacher KD (top-$K$ by gradient alignment) & $0.707$ \\
parameter & continuous push along aligned update directions        & $0.635$ \\
feature   & relational distillation on batch similarity matrices   & $0.611$ \\
feature   & learnable projected bridge ($L = \|h_S - P(h_T)\|^2$)  & $0.624$ \\
\midrule
--- & plain cosine KD baseline & $0.724$ \\
\bottomrule
\end{tabular}
\caption{Transfer channels through distillation.}
\label{tab:kdchannels}
\end{table}

A control worth recording: under the projected-bridge framework,
a \emph{random} teacher scores $+0.014$ \emph{above} the real
teacher --- the spectral-regularisation signature, confirming the
bridge's gain (where any) is regularisation, not transfer.

\subsubsection{Teacher choice in the stream}\label{app:kdteacher}

Three-task stream numglue\_cm $\to$ cstance $\to$ numglue\_ds
(target, $N{=}50$, $50/50$ selected replay of cm; $n{=}5$).
Arms differ only in the KD teacher.  ``Forgetting'' is accuracy
change from the pre-target reference.

\begin{table}[h]
\centering\small
\begin{tabular}{lccc}
\toprule
teacher & target acc & cm forget & cstance forget \\
\midrule
none & $0.151$ & $+0.013$ & $-0.041$ \\
recent (pre-task; default) & $\mathbf{0.167}$ & $0.000$ & $-0.037$ \\
matched (selected source's era) & $0.163$ & $+0.004$ & $-0.043$ \\
base (initial adapter) & $0.151$ & $-0.017$ & $-0.068$ \\
hybrid (matched on replay / recent on target) & $0.163$ & $-0.009$ & $\mathbf{-0.024}$ \\
\bottomrule
\end{tabular}
\caption{Teacher choice in the three-task stream.}
\label{tab:kdstreamteacher}
\end{table}

The replayed task (cm) is protected in every arm --- replay data
itself does that work; in this \emph{short} stream the recent
snapshot is best or tied.  \textbf{In the full eight-task
stream the ranking reverses}: with seven candidate snapshots and
substantial drift accumulated in the recent one, matched-era
teachers win (TRACE-0.5B, $n{=}5$: matched
$0.458 {\scriptstyle\pm .028}$ $>$ random
$0.445 {\scriptstyle\pm .024}$ $>$ recent
$0.439 {\scriptstyle\pm .033}$).  The regime dependence is
mechanistic: the recent snapshot is the most complete teacher
but also the most drifted; as the stream lengthens, drift costs
more than completeness buys.  This motivates the routed-triple
design of \S\ref{sec:method}.

\subsubsection{Lineage, novelty, and mechanism}\label{app:kd:role}

\paragraph{Lineage.}  Distillation for stability in continual
learning descends from LwF~\citep{li2017lwf} (previous model as
teacher, matched on new-task inputs; CNN-era task-incremental
classification) and iCaRL~\citep{rebuffi2017icarl} (exemplars
plus old-class distillation).  DER++~\citep{buzzega2020derpp}
is the closest relative: it stores each buffered sample's
logits at insertion time and anchors to them on replay --- a
frozen, per-sample form of era anchoring.  Our era snapshots
generalise this in three ways: the anchor is a queryable model
rather than stored values (it can answer on any batch,
including current-task batches, and is taken at task end rather
than at insertion); the teacher is bound to the replay data in
one routable record, so a single decision returns both; and the
role assignment is inverted --- DER++ points replay at
stability, we point replay at transfer and give stability to
distillation alone.  Under our protocol the two are also
empirically separated (Table~\ref{tab:method}).  In recent
continual learning with LLMs, distillation appears mainly as
self-distillation~\citep{yang2024sdft}; we are not aware of a
system using routed, era-matched distillation as the explicit
stability module.

\paragraph{Mechanism: why anchoring, not knowledge, in this
regime.}  Three properties of the regime jointly explain the
audit.  (i)~Ground-truth supervision is present on every batch,
so teacher logits carry little marginal label information ---
unlike compression distillation, where the teacher is the
information source.  (ii)~All teachers are small low-rank
perturbations of one frozen base ($r{=}16$, $500$ steps,
$N{=}50$), so every anchor's pull is dominated by the shared
base direction --- which is why identity is second-order (E1,
E4) and why deleting the shared component is catastrophic
(E2).  (iii)~The teacher-specific component is real but
\emph{transient}: it is locally measurable (E5, $r{=}0.93$)
yet, like the one-shot parameter states of
\S\ref{sec:paramfails:carrier}, it does not survive continued
optimisation --- what persists is what is re-injected
identically at every step: the data, and the shared anchor.
There is also a structural asymmetry: replay changes the
\emph{input} distribution, while distillation only tilts the
target distribution on inputs already present --- and a
teacher's source-task knowledge barely manifests on other
tasks' inputs.  Cross-task distillation does carry knowledge
where a jointly trained shared encoder makes the teacher signal
persistent and large (multi-task vision and recommendation
systems); in the carrier taxonomy of
Appendix~\ref{app:scope} those systems occupy the
persistent-parameter cell.  Our negative result is scoped to
the sequential, frozen-base, low-rank-snapshot regime, where we
recommend assigning distillation to stability and spending the
complexity budget on data selection instead.

\section{Experimental Details}\label{app:details}

\subsection{Full-stream setup}\label{app:setupdetail}

\paragraph{Configuration.}  All methods train a single LoRA
adapter ($r{=}16$, $\alpha{=}32$ on the $q,k,v,o$ projections;
AdamW, lr $10^{-4}$, batch size $4$ at 0.5B / $2$ at 7B, max
length $384$) for $500$ steps per task on the $N{=}50$ target
examples, with a replay buffer of $100$ examples per completed
task ($N_{\max}{=}500$ in the \S\ref{sec:transfer} audits).
\methodshort{}: $m{=}10$ probe batches, routing temperature
$\tau{=}0.1$, replay ratio $\alpha{=}0.5$, distillation weight
$\lambda{=}0.5$ at temperature $2$.  The scale audit of
\S\ref{sec:experiments:regime} uses $n{=}5$ seeds per scale
(per-point seed s.e.m.\ $.002$--$.010$).

\paragraph{Evaluation.}  Evaluation uses $500$-example
per-task test sets; classification tasks report exact-match
accuracy, generation tasks evaluation loss.

\paragraph{Stream construction.}  NumGLUE-8 arranges the eight
task types of the established NumGLUE
suite~\citep{mishra2022numglue} as a continual stream --- the
same construction by which Split-CIFAR derives from CIFAR and
TRACE itself from eight existing datasets; two of its types
(numglue-cm/ds) already appear inside TRACE.

\paragraph{Cost of \methodshort{}.}  Signatures are $m{=}10$
forward-backward passes per task (all eight of a benchmark's
signatures take ${\sim}30$ seconds at 0.5B, and picks are
scale-stable, so a small proxy model suffices); routing is an
inner product; training cost matches standard replay; the
buffer stores $100$ examples per task (${\sim}3$\,MB for $15$
tasks, versus ${\sim}130$\,MB for per-task adapter libraries).
We regard this routable task record --- signature, data
subsample, snapshot --- rather than the example-level buffer of
rehearsal methods or the adapter-only library of
parameter-isolation methods, as the right unit of continual
memory.

\paragraph{New-baseline ports.}  GainLoRA and DEAL
(NeurIPS'25) are ported from their official releases into the
stream protocol, mechanism-specific components kept as released
(GainLoRA's projection-constrained model-level gate; DEAL's
Haar-DWT retention filter) and only the adapter footprint
unified for parity ($q,k,v,o$ at $r{=}16$, $\alpha{=}32$, lr
$10^{-4}$).  Two adaptations: DEAL's per-task re-initialisation
depends on a checkpoint-namespace quirk that cannot restore the
adapter from task~3 on, so we adopt the continuous reading
consistent with the paper's framing; and at DEAL's native lr
$10^{-5}$ the $500$-step budget under-trains in all four
settings (TRACE-0.5B overall $.068$ vs.\ $.372$), so the
standardised rate is also the charitable one.

\subsection{Per-task results}\label{app:pertask}

End-of-stream accuracy per classification task, TRACE-8 at 0.5B
(canonical order, mean over $5$ seeds; the per-task view behind
Table~\ref{tab:method}).

\begin{table}[h]
\centering\scriptsize
\setlength{\tabcolsep}{3pt}
\resizebox{\textwidth}{!}{%
\begin{tabular}{l cccccccccccc}
\toprule
Task & SeqFT & EWC & LwF & SDFT & A-GEM & O-LoRA & DER++ & ER &
\methodshort{}\textsubscript{arg} & \methodshort{}\textsubscript{soft} &
\methodshort{}\textsubscript{hyb} & \methodshort{}\textsubscript{tri} \\
\midrule
C-STANCE   & $.273$ & $.178$ & $.303$ & $.260$ & $.265$ & $.290$ & $.456$ & $.463$ & $.481$ & $.522$ & $.510$ & $.521$ \\
FOMC       & $.214$ & $.215$ & $.167$ & $.202$ & $.390$ & $.178$ & $.075$ & $.580$ & $.502$ & $.553$ & $.536$ & $.541$ \\
ScienceQA  & $.491$ & $.473$ & $.533$ & $.538$ & $.533$ & $.426$ & $.000$ & $.593$ & $.579$ & $.567$ & $.612$ & $.600$ \\
NumGLUE-cm & $.225$ & $.255$ & $.320$ & $.383$ & $.262$ & $.188$ & $.411$ & $.474$ & $.483$ & $.506$ & $.485$ & $\mathbf{.552}$ \\
NumGLUE-ds & $.117$ & $.107$ & $.132$ & $.144$ & $.102$ & $.084$ & $.157$ & $.147$ & $.152$ & $.147$ & $.151$ & $\mathbf{.159}$ \\
\midrule
\textbf{AVG} & $.264$ & $.245$ & $.291$ & $.305$ & $.310$ & $.233$ & $.220$ & $.451$ & $.439$ & $.459$ & $.459$ & $\mathbf{.475}$ \\
\bottomrule
\end{tabular}}%

\caption{Per-task end-of-stream accuracy on TRACE-8 at 0.5B.}
\label{tab:pertaskfull}
\end{table}

Per-task tables for the remaining settings follow the same
pattern and are available with the released evaluation matrices.

\subsection{Stream robustness}\label{app:robust}

\paragraph{Task order.}  Across the greedy best-source
curriculum, its reversal, two deterministic reshufflings, and
arbitrary reorderings, at both scales, the replay family varies
by at most $.03$ in ranking and magnitude; no-memory baselines
vary substantially, tracking the order's \emph{tail} ---
recency, not the matrix: at 0.5B they swing by $+.05$ to $+.09$
under the curriculum, and an arbitrary reshuffle with a similar
tail lifts SeqFT even more ($+.12$); at 7B, where the base
barely forgets, the same swings shrink below $.04$, and on
NumGLUE all order effects are within seed noise
(\S\ref{sec:experiments:ablations}).
The one mild sensitivity is argmax selection when a target
precedes its best sources --- the transfer matrix is
directional, and soft routing removes this by construction.

\paragraph{Backbone.}  We further evaluate whether the
framework generalises across backbone families.  Following
\S\ref{sec:transfer}, we first audit headroom: on
Llama-3.2-1B-Instruct~\citep{llama3} and
Gemma-3-1B-it~\citep{gemma3}, the TRACE-8 audit finds open
headroom on every task (target-full minus target-data, $n{=}2$:
$+.045$ to $+.135$ accuracy on Llama and $+.045$ to $+.269$ on
Gemma across the five classification targets; the three
generation targets improve by $1.8$--$4.6$ in loss), so
transfer is measurable on both.
Table~\ref{tab:backbonefull} reports the stream under the
exact configuration of Appendix~\ref{app:setupdetail}.
The qualitative picture of Table~\ref{tab:method} is
unchanged: replay-based methods dominate; \methodshort{}
attains the best overall accuracy, plasticity, and backward
transfer on both backbones; and positive backward transfer
remains confined to the replay-based methods --- every
rehearsal-free baseline, including the expansion-based
GainLoRA, forgets.

\begin{table}[h]
\centering\small
\setlength{\tabcolsep}{3.5pt}
\begin{tabular}{l ccc ccc}
\toprule
 & \multicolumn{3}{c}{TRACE-8 (Llama-3.2-1B)} & \multicolumn{3}{c}{TRACE-8 (Gemma-3-1B)} \\
\cmidrule(lr){2-4}\cmidrule(lr){5-7}
Method & Overall & Plas & BWT & Overall & Plas & BWT \\
\midrule
SeqFT
 & $.420{\scriptscriptstyle\pm .028}$ & $.458{\scriptscriptstyle\pm .018}$ & $-.038{\scriptscriptstyle\pm .022}$ & $.154{\scriptscriptstyle\pm .018}$ & $.319{\scriptscriptstyle\pm .012}$ & $-.164{\scriptscriptstyle\pm .023}$ \\
\midrule
LwF
 & $.422{\scriptscriptstyle\pm .027}$ & $.459{\scriptscriptstyle\pm .014}$ & $-.038{\scriptscriptstyle\pm .015}$ & $.241{\scriptscriptstyle\pm .028}$ & $.319{\scriptscriptstyle\pm .012}$ & $-.078{\scriptscriptstyle\pm .020}$ \\
GainLoRA
 & $.430{\scriptscriptstyle\pm .019}$ & $.438{\scriptscriptstyle\pm .019}$ & $-.008{\scriptscriptstyle\pm .009}$ & $.279{\scriptscriptstyle\pm .013}$ & $.301{\scriptscriptstyle\pm .010}$ & $-.022{\scriptscriptstyle\pm .012}$ \\
\midrule
ER
 & $.494{\scriptscriptstyle\pm .019}$ & $.470{\scriptscriptstyle\pm .012}$ & $+.024{\scriptscriptstyle\pm .020}$ & $.354{\scriptscriptstyle\pm .011}$ & $.337{\scriptscriptstyle\pm .012}$ & $+.018{\scriptscriptstyle\pm .017}$ \\
\midrule
\textbf{\methodshort{}}
 & $\mathbf{.519}{\scriptscriptstyle\pm .015}$ & $\mathbf{.481}{\scriptscriptstyle\pm .012}$ & $\mathbf{+.037}{\scriptscriptstyle\pm .011}$ & $\mathbf{.381}{\scriptscriptstyle\pm .014}$ & $\mathbf{.351}{\scriptscriptstyle\pm .012}$ & $\mathbf{+.030}{\scriptscriptstyle\pm .010}$ \\
\bottomrule
\end{tabular}
\caption{\textbf{Backbone robustness of \methodshort{} on
TRACE-8} under the standard sequential protocol (Llama-3.2-1B
and Gemma-3-1B; mean $\pm$ std over 5 seeds).}
\label{tab:backbonefull}
\end{table}

\subsection{Reproducibility}\label{app:repro}

Base models: Qwen2.5-0.5B-Instruct and Qwen2.5-7B-Instruct
\citep{qwen25}; the backbone-robustness check additionally
uses Llama-3.2-1B-Instruct~\citep{llama3} and
Gemma-3-1B-it~\citep{gemma3}.  Training configuration as in
Appendix~\ref{app:setupdetail} (PP15 runs use max length
$256$).  Probe protocol, arm definitions, and per-cell seed
counts as stated inline; full-seed statistics are reported
throughout to avoid favourable-subset selection
bias~\citep{li2026personalized}.  A minimal implementation reproducing the
core TRACE-8 experiments is released at
\url{https://github.com/ymeng3/transfer-selective-replay}.

\end{document}